\documentclass[10pt,twocolumn,letterpaper]{article}

\usepackage[margin=0.9in]{geometry}
\usepackage{times}
\usepackage{graphicx}
\usepackage{amsmath}
\usepackage{amssymb}

\usepackage{subcaption}
\usepackage{multirow}
\usepackage[linesnumbered,ruled,vlined]{algorithm2e}

\newcommand{\sqr}{\mathsmaller{\square}}
\newcommand{\dmd}{\mathlarger{\diamond}}

\graphicspath{{figures/}}

\usepackage[pagebackref=true,breaklinks=true,letterpaper=true,colorlinks,bookmarks=false]{hyperref}

\begin{document}

\title{Fast Screening Algorithm for Rotation and Scale Invariant Template Matching}

\author{Bolin Liu\\
McMaster University\\
{\tt\small liub30@mcmaster.ca}
\and
Xiao Shu\\
McMaster University\\
{\tt\small shux@mcmaster.ca}
\and
Xiaolin Wu\\
Shanghai Jiao Tong University\\
{\tt\small xwu510@sjtu.edu.cn}
}

\date{}

\maketitle
\thispagestyle{empty}

\begin{abstract}
  This paper presents a generic pre-processor for expediting
  conventional template matching techniques.  Instead of locating the
  best matched patch in the reference image to a query template via
  exhaustive search, the proposed algorithm rules out regions with no
  possible matches with minimum computational efforts.  While working
  on simple patch features, such as mean, variance and gradient, the
  fast pre-screening is highly discriminative. Its computational
  efficiency is gained by using a novel octagonal-star-shaped template
  and the inclusion-exclusion principle to extract and compare patch
  features.  Moreover, it can handle arbitrary rotation and scaling of
  reference images effectively.  Extensive experiments demonstrate
  that the proposed algorithm greatly reduces the search space while
  never missing the best match.
\end{abstract}

\section{Introduction}
\label{sec:introduction}

Template matching, a fundamental operation in computer vision, is to
locate the best matched patch in a reference image to a given query
template.  While finding an object in image seems to be trivial for
human in general, it is a surprisingly challenging problem for
computer algorithms.  Most existing template matching techniques are
very time-consuming and demand a lot of computational resources to
obtain relatively reliable results in real-world applications, where
geometric distortions between the reference image and query template
are unavoidable.

To handle geometric distortions like a simple change in scale or
orientation, conventional patch based template matching techniques,
which compare the template directly with all the candidate patches in
succession, have to rely on an exhaustive search of all the
combinations of different scales and rotations.  In comparison, newly
emerged feature based techniques are more efficient and robust for
matching a template with deformations.  The basic idea of feature
based template matching is to extract some statistical features from
the query template and check if these features also occur in some
patches in the reference image.  As matched features do not need to
appear at the exact matched positions, feature based techniques are
more resilient to geometric distortions as long as both of the
reference image and query template have distinctive features.
However, with regard to overall matching speed, feature based
techniques are still unsatisfactory; they commonly require a
computationally intensive process to generate features for different
scales, and the whole matching process can be slow for high-resolution
images.

In this paper, we propose a scale and rotation invariant template
matching pre-processor for expediting conventional patch and feature
based template matching techniques.  Instead of pinpointing the best
matched patch, this pre-processor rapidly rules out regions of no
possible matches using simple patch features, such as mean, variance
and gradient.  After the pre-processing procedure, the remainder parts
of the reference image are passed to a more accurate but slower
template matching technique to locate the exact position of the best
match.  As the regions to search are greatly reduced in the
pre-processing step, the accurate template matching technique needs to
process far fewer pixels or patches hence it runs much faster than
having to process the whole image.

To make this two-stage method a viable strategy, the cost of the
pre-processing, i.e., the time spent on identifying no match regions,
must be less than the time saved in the later accurate template
matching stage.  Thus, the major challenge in the design of the
pre-processing algorithm is to make its computational complexity
extremely low while effective enough to mark as many unmatched regions
as possible.  On the other hand, since the accurate template matching
technique in the second stage only operates on the unmarked regions
from the first stage, if the best matched patch has already been
incorrectly marked as unmatched, the two-stage method would fail to
locate the best match.  Therefore, in order to avoid affecting the
success rate of the subsequent template matching, the false positive
rate of the pre-processing algorithm must be made very low.

The remainder of this paper is organized as follows.  Section
\ref{sec:related} reviews some of the related template matching
techniques.  Section \ref{sec:feature} discusses the patch features
employed by the proposed technique.  Sections \ref{sec:rotation} and
\ref{sec:scale} elaborate on how to make the proposed technique
rotation and scale invariant, respectively.  Section~\ref{sec:result}
presents experimental results and Section~\ref{sec:conclusion}
concludes.

\section{Related Work}
\label{sec:related}

A great amount of research effort has been dedicated to designing
efficient and effective template matching techniques.  Based on
conventional full search matching, Alkhansari proposed a technique
that reduces search space by pruning unmatched regions using a
downsampled reference image \cite{gharavi2001fast}.  Pele and Werman
developed a method to determine the optimal step size of sliding
windows for full search matching \cite{pele2007accelerating}.  On the
topic of latest full search equivalent techniques, Ouyang \textit{et
  al.} provided a comprehensive survey and compared the performances
of several popular methods \cite{ouyang2012performance}.

Template matching techniques based on SSD or normalized cross
correlation (NCC) can be accelerated using frequency domain approaches
\cite{brown1992survey} or summer area tables \cite{lewis1995fast,
  crow1984summed, viola2004robust}.  SAD, SSD and NCC based techniques
are often highly efficient in terms of computational cost, however,
they are not flexible enough to handle geometric transformation
between the reference image and query template efficiently.  Ullah and
Kaneko used the orientation code to represent gradient information in
a patch for approximating rotation angles as well as for rotation
invariant matching \cite{ullah2004using}. Choi and Kim proposed to
combine circular projection and Zernike moments to achieve rotation
invariance \cite{choi2002novel}.  Their method was later improved by
Kim \cite{kim2010rotation}.  Based on circular projection, Kim used
the Fourier coefficients of radial projections as rotation invariant
features.  Lin and Chen used ring projection transform to establish
parametric template vector for differently scaled template to get
invariance for rotation and scale \cite{lin2008template}.  Moreover,
Tsai and Chiang \cite{tsai2002rotation} solved the rotation problem
using wavelet decompositions and ring projection.  Korman \textit{et
  al.} \cite{korman2013fast} approximated the 2D affine transformation
in reference image by constructing a transformation net with SAD error
level $\delta$.  Wakahara and Yamashita \cite{wakahara2014gpt}
proposed a global projection transform correlation method to deal with
arbitrary 2D transformation.

Recently, feature-based image matching methods, such as scale
invariant feature transform (SIFT) \cite{lowe2004distinctive}, become
more popular.  After obtaining rotation and scale invariant features
for both template and reference image, data fitting algorithm like
RANSAC \cite{fischler1981random} are used for finding matching
patterns.  ASIFT extends SIFT to be fully affine invariant
\cite{morel2009asift}.  Dekel \textit{et al} introduced a novel
similarity measure termed best-buddies similarity (BBS) for comparing
the features of two patches \cite{dekel2015best}.  BBS is robust
against many types of geometric deformations and well suited for video
tracking applications.  These feature-based methods are generally time
consuming due to their heavy processes for generating feature
descriptors, also they may fail to work if the template is relatively
small or lightly textured.

\section{Patch Features}
\label{sec:feature}

The proposed algorithm rules out regions with no possible matches
based on whether or not the patches in these regions share the same
features with the given query template.  Patch features employed by
conventional template matching techniques are often complex and
computationally expensive due to the accuracy and robustness
requirements for pinpointing the best match.  However, since the
proposed pre-processing algorithm only needs to rule out patches that
are significantly different from the query template, simple patch
features, such as mean, variance and gradient, are sufficient for
distinguishing a majority of those unmatched patches.

\begin{figure}
  \centering
  \includegraphics[width=0.8\linewidth]{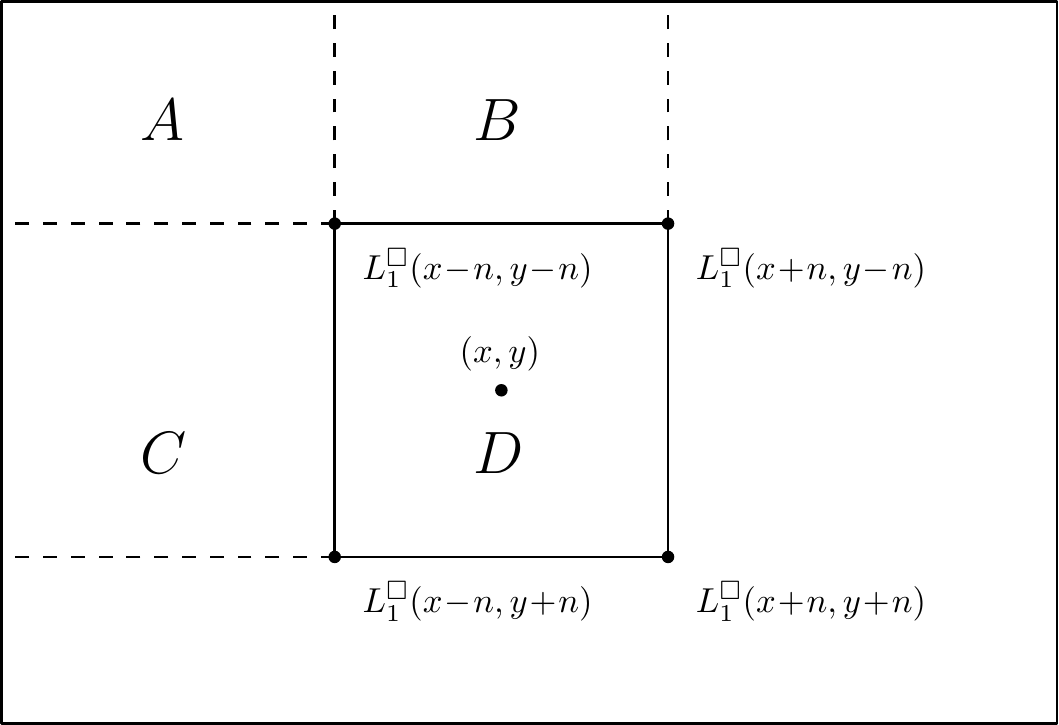}
  \caption{The sum $S^{\sqr}_1(x,y)$ of area $D$ is a simple linear
    combination of $L^{\sqr}_1(x\!+\!n, y\!+\!n),
    L^{\sqr}_1(x\!-\!n, y\!+\!n), L^{\sqr}_1(x\!+\!n, y\!-\!n),
    L^{\sqr}_1(x\!-\!n, y\!-\!n)$, which represent the sums of area
    $A+B+C+D$, $A+C$, $A+B$ and $A$, respectively}
  \label{fig:sqr_mean}
\end{figure}

The mean of a patch, the inner product of the patch with a box average
kernel, can be calculated efficiently by using the inclusion-exclusion
principle.  As presented in \cite{crow1984summed}, it only needs one
addition and two subtractions to calculate the sum of the pixel
intensity of an arbitrary rectangular area using a summed area table,
which is also known as integral image.  The idea is that, if we know
the sum $L^{\sqr}_1(x,y)$ of a rectangular area with top-left
corner $(1,1)$ and bottom-right corner $(x,y)$ for any given pixel
$(x,y)$ of image $I$, as shown in Figure~\ref{fig:sqr_mean}, then we
can calculate the sum $S^{\sqr}_1(x,y)$ of any $2n \times 2n$ patch
centered at $(x,y)$ as follows,
\begin{align}
  S^{\sqr}_1(x,y)
  = & \sum_{i=x-n+1}^{x+n}
    \sum_{j=y-n+1}^{y+n} I(i,j) \nonumber \\
  = & L^{\sqr}_1(x\!+\!n, y\!+\!n)
    - L^{\sqr}_1(x\!-\!n, y\!+\!n) \nonumber \\
  & - L^{\sqr}_1(x\!+\!n, y\!-\!n) + L^{\sqr}_1(x\!-\!n, y\!-\!n),
  \label{eq:S_1}
\end{align}
where the summed area table $L^{\sqr}_1(x,y)$, by its definition,
is,
\begin{align}
  L^{\sqr}_1(x,y) = \sum_{i=1}^x \sum_{j=1}^y I(i,j).
  \label{eq:T_1}
\end{align}
Given the sum of a patch centered at $(x,y)$, the mean
$S^{\sqr}_{\mu}(x,y)$ of the patch is,
\begin{align}
  S^{\sqr}_{\mu}(x,y) = S^{\sqr}_1(x,y) / (2n)^2.
  \label{eq:S_mu}
\end{align}
The summed area table $L^{\sqr}_1(x,y)$ for image $I$ can be
constructed efficiently by using a pass of horizontal cumulative sum
on every row of $I$ followed by a pass of vertical cumulative sum on
every column.  Since the cumulative sums of different rows (or
columns) are independent, it is easy to accelerate the construction
process with parallel computing architecture, such as GPU, by
calculating the cumulative sums of rows (or columns) concurrently.
After the linear-time construction of the summer area table
$L^{\sqr}_1(x,y)$, each of the following query for the intensity
sum $S^{\sqr}_1(x,y)$ or mean $S^{\sqr}_{\mu}(x,y)$ of an arbitrary
patch only requires constant time regardless of the size the patch.

\begin{figure}
  \centering
  \begin{subfigure}{0.45\linewidth}
    \centering
    \includegraphics[width=0.8\textwidth]{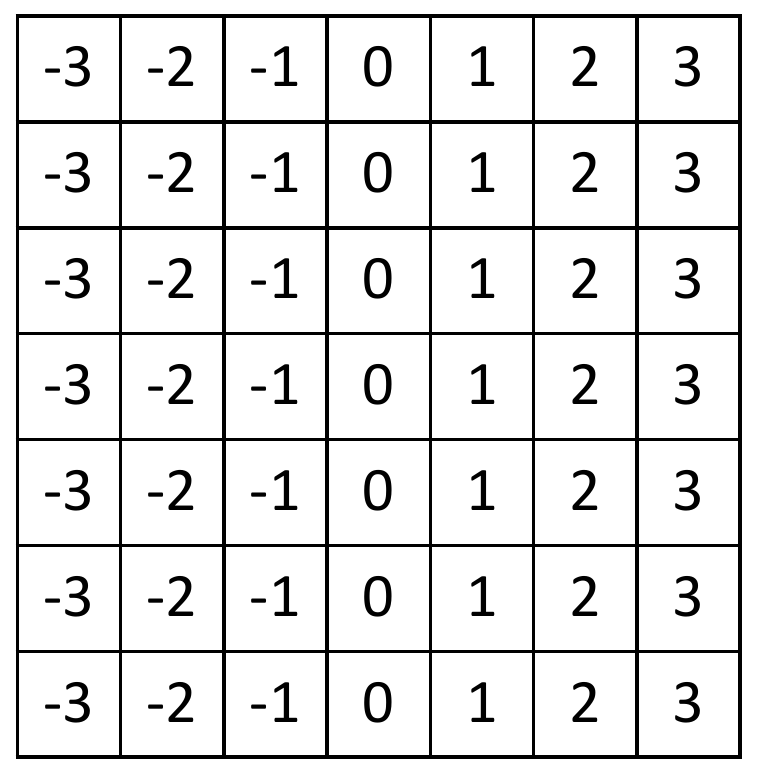}
    \caption{Horizontal}
  \end{subfigure}
  \begin{subfigure}{0.45\linewidth}
    \centering
    \includegraphics[width=0.8\textwidth]{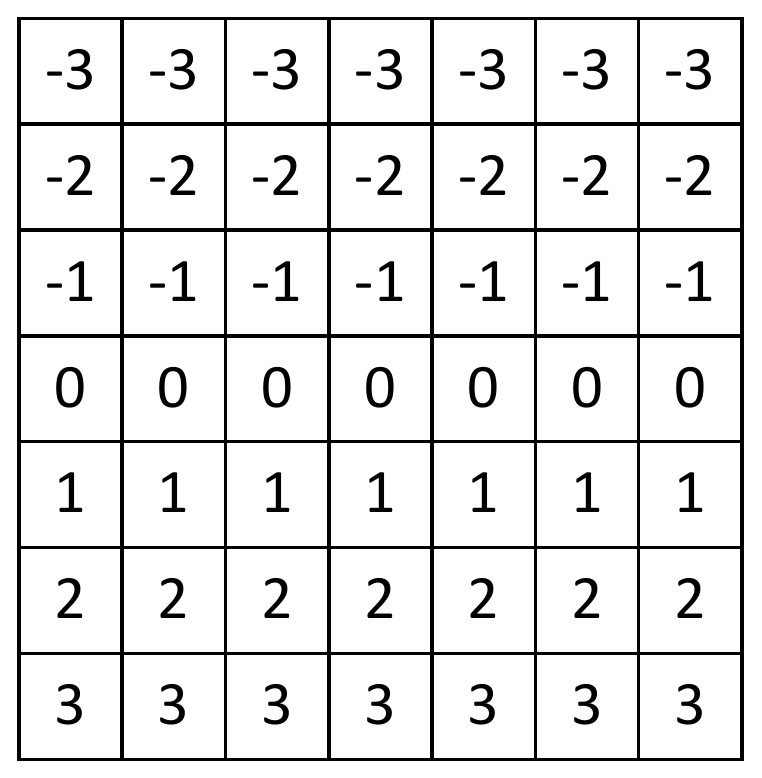}
    \caption{Vertical}
  \end{subfigure}
  \caption{The extended $7 \times 7$ Prewitt kernels, which are
    basically linear gradient ramps.}
  \label{fig:prewitt}
\end{figure}

In addition to the mean of a patch, the proposed algorithm also
employs two other linear features, the horizontal and vertical
gradients.  The horizontal and vertical gradient of a patch,
representing the rate of pixel intensity change of the patch from left
to right and from top to bottom respectively, are the inner products
of the patch with the extended Prewitt kernels \cite{kekre2010image}
of the same size.  The extended Prewitt kernels are basically linear
gradient ramps as shown in Figure~\ref{fig:prewitt}, and they are
defined as follows,
\begin{align}
  P_{x}(x,y) &= x-\frac{m+1}{2}, \nonumber \\
  P_{y}(x,y) &= y-\frac{m+1}{2},
\end{align}

\begin{figure}
  \centering

  \begin{subfigure}{0.45\linewidth}
    \centering
    \includegraphics[width=0.8\textwidth]{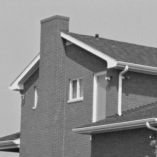}
    \caption{Input image}
    \label{fig:pca:img}
  \end{subfigure}
  \begin{subfigure}{0.45\linewidth}
    \centering
    \includegraphics[width=0.8\textwidth]{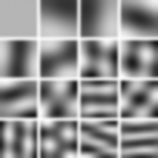}
    \caption{First 16 components}
    \label{fig:pca:16}
  \end{subfigure}

  \caption{A sample image and the first 16 principle components of all
    the $16 \times 16$ patches in the image
    \cite{deledalle2011image}.}
  \label{fig:pca}
\end{figure}

Choosing horizontal and vertical gradients as patch features is based
on the fact that the first three principle components of natural image
patches in principle component analysis (PCA) are the box average
kernel, vertical extended Prewitt kernel and horizontal extended
Prewitt kernel, respectively (as shown in Figure~\ref{fig:pca}).
Thus, for distinguishing patches, these kernels have the strongest
discriminating power among all the combinations of any three linear
features.

In a similar fashion as the sum of a patch as in Eq.~\ref{eq:S_1},
gradient of a patch can also be calculated efficiently by using the
inclusion-exclusion principle.  For instance, the horizontal gradient
$S^{\sqr}_x(x,y)$ of a $2n \times 2n$ patch centered at $(x,y)$ is,
\begin{align}
  S^{\sqr}_x(x,y)
  = & \sum_{i=x-n+1}^{x+n}
    \sum_{j=y-n+1}^{y+n} (i-x) \cdot I(i,j) \nonumber \\
  = & L^{\sqr}_x(x\!+\!n, y\!+\!n)
    - L^{\sqr}_x(x\!-\!n, y\!+\!n) \nonumber \\
  & - L^{\sqr}_x(x\!+\!n, y\!-\!n) + L^{\sqr}_x(x\!-\!n, y\!-\!n) \nonumber \\
  & - x S^{\sqr}_1(x,y)
  \label{eq:S_y}
\end{align}
where $L^{\sqr}_x(x,y)$, the summed area map of image $i \cdot
I(i,j)$, is,
\begin{align}
  L^{\sqr}_x(x,y) &= \sum_{i=1}^x \sum_{j=1}^y i \cdot I(i,j).
  \label{eq:T_x}
\end{align}
Similarly, the vertical gradient $S^{\sqr}_y(x,y)$ of the patch is
tractable with the same technique.

Another patch feature employed by the proposed algorithm is variance.
Given the mean $S^{\sqr}_{\mu}(x,y)$ and second moment
$S^{\sqr}_2(x,y)$ of a $2n \times 2n$ patch, the variance
$S^{\sqr}_{\sigma}(x,y)$ of the patch can be obtained as,
\begin{align}
  S^{\sqr}_{\sigma}(x,y) = S^{\sqr}_2(x,y)/(2n)^2-[S^{\sqr}_{\mu}(x,y)]^2.
  \label{eq:S_sigma}
\end{align}
Since the second moment $S^{\sqr}_2(x,y)$ of each patch,
\begin{align}
  S^{\sqr}_2(x,y) = \sum_{i=x-n+1}^{x+n}
    \sum_{j=y-n+1}^{y+n} [I(i,j)]^2,
\end{align}
only requires constant time to calculate with the summed area table of
$[I(i,j)]^2$, we can acquire the variance of an arbitrary patch
efficiently as well.

\section{Rotation Invariance}
\label{sec:rotation}

The patch features discussed in the previous section are excellent
indicators for ruling out unmatched patches if the object in the query
template and reference image shares the same orientation and scale.
However, if that is not the case, a small mismatch between the
orientations of the query template and reference image could result
distinct patch features for the same object, causing mistakes in
matching the object.  This over-sensitivity to orientation is due to
two problems: first, some of patch features like horizontal and
vertical gradients are not rotation invariant; second, the shape of
the template is also not rotation invariant.

The first problem can be easily resolved by replacing the horizontal
and vertical gradients with the magnitude of gradient,
\begin{align}
  S^{\sqr}_m(x,y) &=\sqrt{[S^{\sqr}_x(x,y)]^2+[S^{\sqr}_y(x,y)]^2},
\end{align}
which is robust against rotation and still simple to calculate.

\begin{figure}[t]
  \centering
  \begin{subfigure}{0.4\linewidth}
    \centering
    \includegraphics[width=0.8\textwidth]{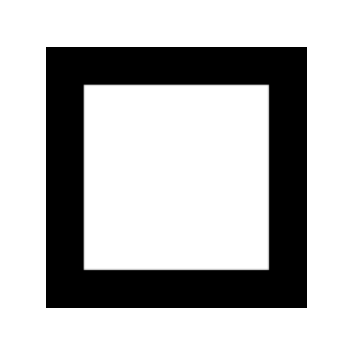}
    \caption{Template}
  \end{subfigure}
  \begin{subfigure}{0.4\linewidth}
    \centering
    \includegraphics[width=0.8\textwidth]{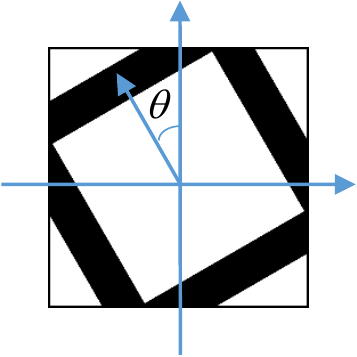}
    \caption{Reference image}
  \end{subfigure}
  \begin{subfigure}{0.3\linewidth}
    \centering
    \includegraphics[width=0.95\textwidth]{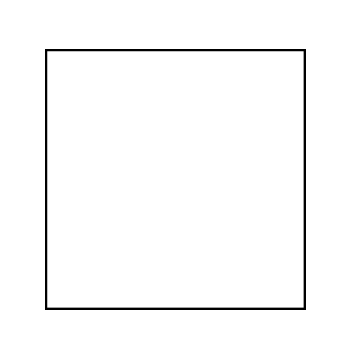}
    \caption{Square}
  \end{subfigure}
  \begin{subfigure}{0.3\linewidth}
    \centering
    \includegraphics[width=0.95\textwidth]{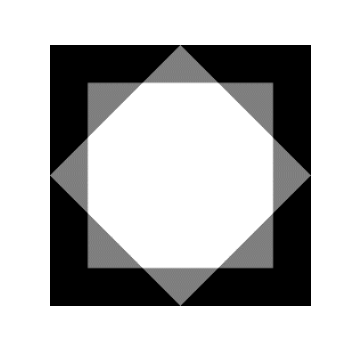}
    \caption{Octagonal star}
  \end{subfigure}
  \begin{subfigure}{0.3\linewidth}
    \centering
    \includegraphics[width=0.95\textwidth]{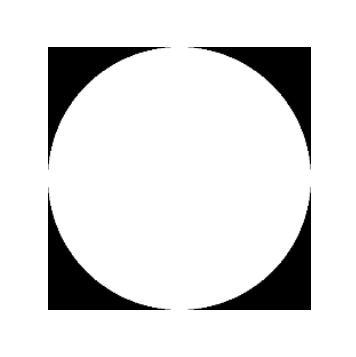}
    \caption{Circle}
  \end{subfigure}
  \begin{subfigure}{0.90\linewidth}
    \includegraphics[width=\textwidth]{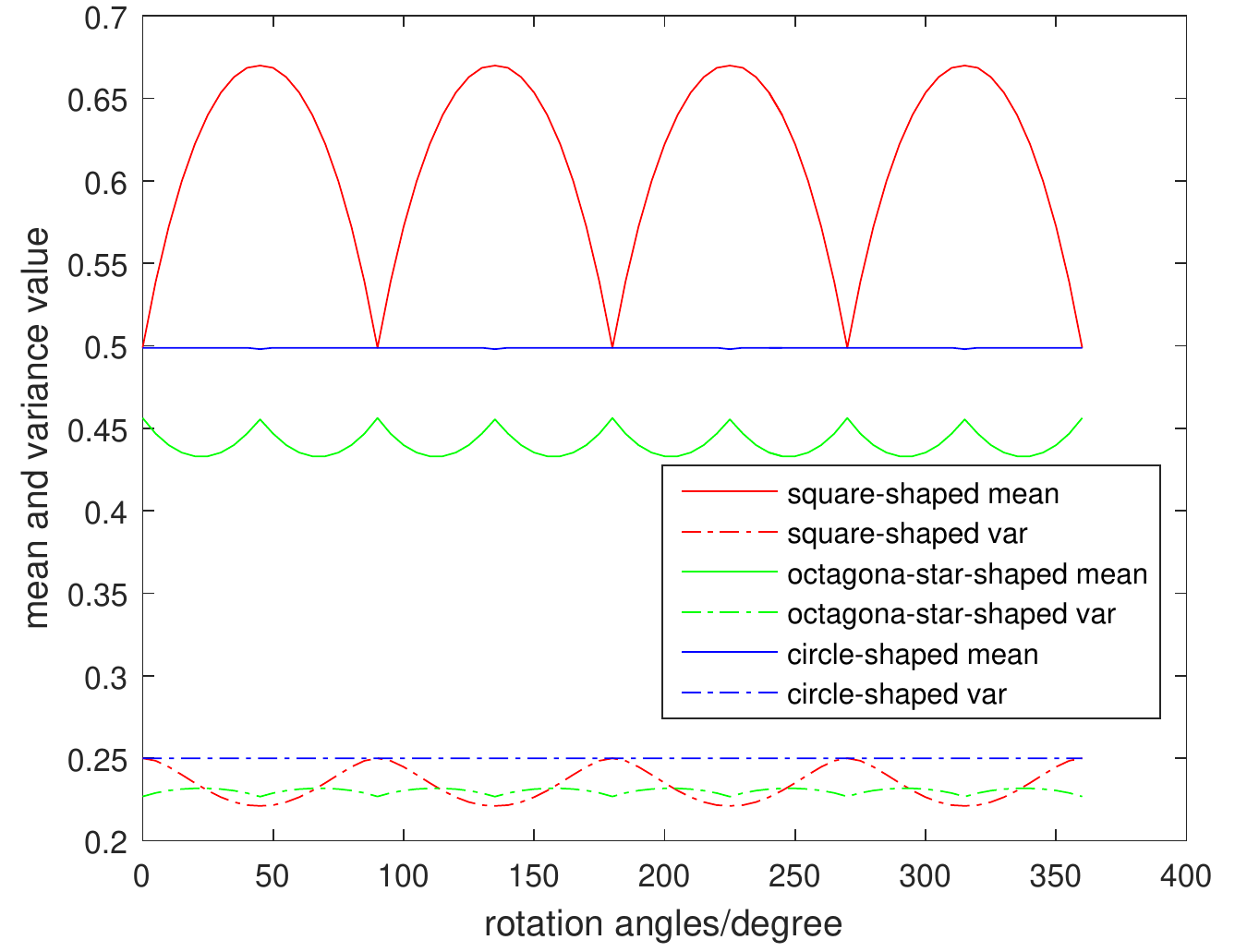}
    \caption{The mean and variance of a patch plotted as a function of
      image rotation angle.}
    \label{fig:sequare:meanvar}
  \end{subfigure}
  \caption{Octagonal-star-shaped template is much more robust against
    rotation than square template.}
  \label{fig:square}
\end{figure}

To solve the second problem, we need to change the shape of the
template.  Commonly, the query template is a square image patch
provided by the user.  Given the template, the pre-processing
algorithm then examines the features of every square patch of the same
size as the template in the reference image.  The drawback of using
square patch is the area covered by a square patch are variant to
image rotation.  For instance, the reference image shown in
Figure~\ref{fig:square} is a black framed box on white background.  A
rotation of the reference image moves some of the white background
pixels into the window centered at the box while cropping out some of
the black frame pixels.  The resulting change in the set of pixels in
the window alters all its patch features such as the mean and variance
as in Figure~\ref{fig:square}.

\begin{figure}
  \centering
  \includegraphics[width=1\linewidth]{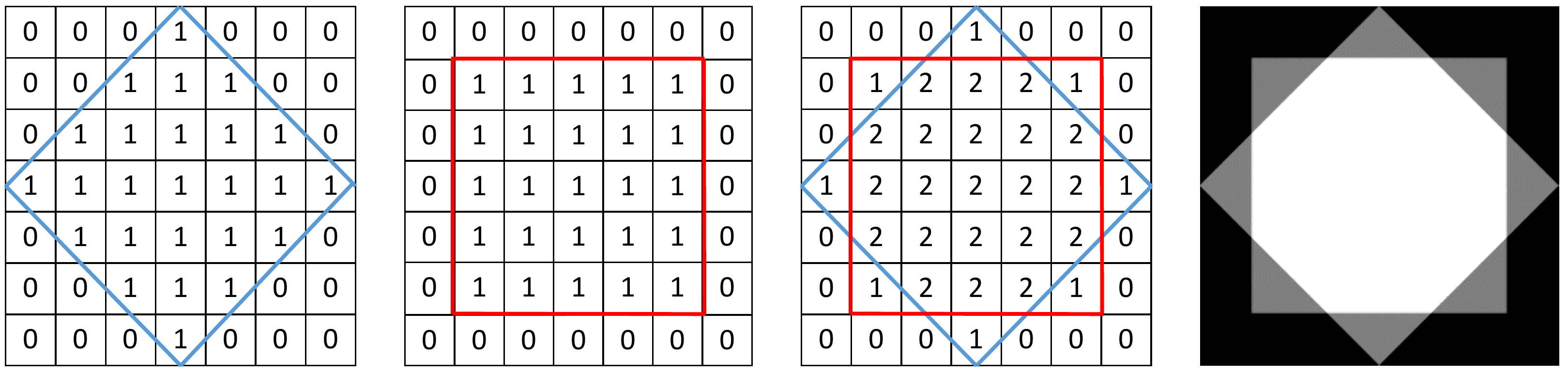}
  \caption{An example of octagonal-star-shaped 7$\times$7 template
    (the third figure), which is the superposition of a square (the
    second figure) and a diamond (the first figure).}
  \label{fig:os}
\end{figure}

Ideally, the best template shape for rotation invariant matching is a
circle, but it is not possible to accelerate the feature calculation
of an arbitrary circular patch using the inclusion-exclusion technique
introduced previously.  Thus, we propose octagonal-star-shaped
template, which approximates to a circle but still has efficient
feature calculation algorithm.  An octagonal star is the superposition
of a square and a diamond (square rotated by $45^{\circ}$ ) of the
same size as depicted in Figure~\ref{fig:os}.  The weighted mean
$S_{\mu}(x,y)$ of an octagonal-star-shaped template, for example, is
the arithmetic average of the mean $S^{\sqr}_{\mu}(x,y)$ of the square
and the mean $S^{\dmd}_{\mu}(x,y)$ of the diamond in the template,
i.e.,
\begin{align}
  S_{\mu}(x, y) = \frac{1}{2}[S^{\sqr}_{\mu}(x,y) + S^{\dmd}_{\mu}(x,y)],
  \label{eq:S_os}
\end{align}

\begin{figure}
  \centering
  \includegraphics[width=0.8\linewidth]{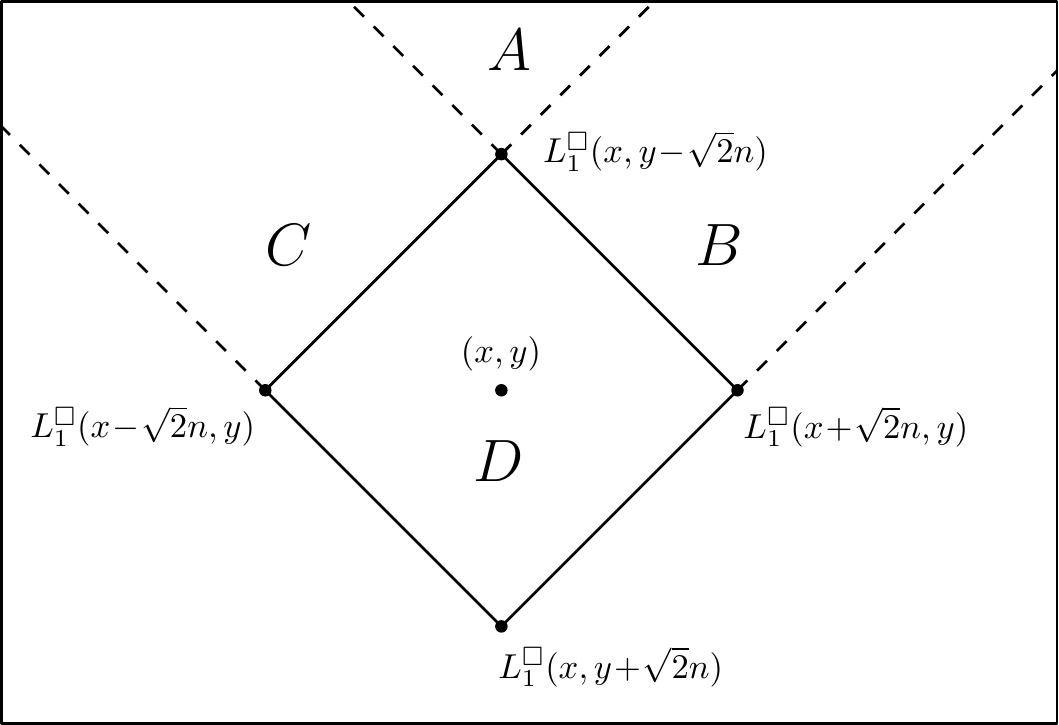}
  \caption{The sum $S^{\dmd}_1(x,y)$ of area $D$ is a simple linear
    combination of $L^{\dmd}_1(x, y\!+\!\sqrt{2} n),
    L^{\dmd}_1(x\!-\!\sqrt{2} n, y), L^{\dmd}_1(x\!+\!\sqrt{2} n, y),
    L^{\dmd}_1(x, y\!-\!\sqrt{2}n)$, which represent the sums of area
    $A+B+C+D$, $A+C$, $A+B$ and $A$, respectively}
  \label{fig:dmd_mean}
\end{figure}

Similar to the previous rectangular case, if the sum $L^{\dmd}_1(x,y)$
of the triangular area with right angle vertex $(x,y)$, as shown in
Figure~\ref{fig:dmd_mean}, is known for any $x,y$, where
$L^{\dmd}_1(x,y)$ is,
\begin{align}
  L^{\dmd}_1(x,y) = \sum_{i=\max(y-x,1)}^{\min(y+x, M)} \sum_{j=1}^{y-|x-i|} I(i,j)
\end{align}
and $M$ is the width of the image.  Then, by the inclusion-exclusion
principle, we can calculate the sum $S^{\dmd}_1(x,y)$ and mean
$S^{\dmd}_{\mu}(x,y)$ of a $2n \times 2n$ diamond area centered at
$(x,y)$ using one addition and two subtractions as follows
\cite{lienhart2002extended, messom2006fast},
\begin{align}
  S^{\dmd}_1(x,y)
  = & L^{\dmd}_1(x, y\!+\!\sqrt{2} n)
      - L^{\dmd}_1(x\!-\!\sqrt{2} n, y) \nonumber \\
  & - L^{\dmd}_1(x\!+\!\sqrt{2} n, y)
    + L^{\dmd}_1(x, y\!-\!\sqrt{2}n) \\
  S^{\dmd}_{\mu}(x,y) = & S^{\dmd}_1(x,y) / (2n)^2
\end{align}
Same as $L^{\sqr}_1(x,y)$, the construction of the summed triangular
area map $L^{\dmd}_1(x,y)$ is tractable in linear time using two
diagonal passes of cumulative sum.  Besides the mean, other patch
features, such as the variance and magnitude of gradient, of a
octagonal-star-shaped template can also be computed efficiently by
combining the features of the square and diamond areas in the
template.



\section{Scale Invariance}
\label{sec:scale}

\begin{figure}
  \centering
  \includegraphics[width=0.95\linewidth]{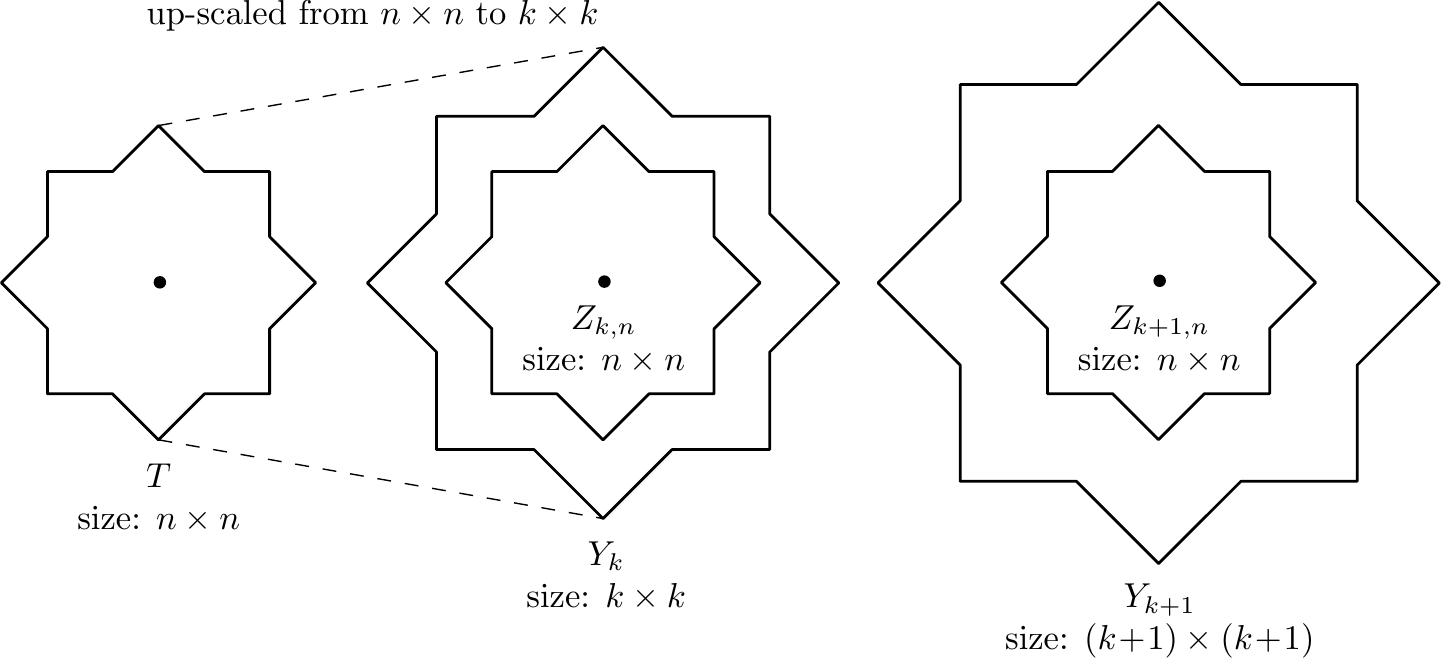}
  \caption{Templates in different scales.}
  \label{fig:scale}
\end{figure}

Another challenge for the proposed pre-processing algorithm is how to
handle the scale difference between the query template and reference
image.  In the previous sections, we assume that the scales of the
query template and reference image are identical, hence the best
matched patch must be of the exact same size as the query template.
However, this assumption is impractical in most real-world
applications; the scale information is usually unknown and the best
matched patch may be larger or smaller than the query template.

One solution to this scale invariant problem is simply to scale the
query octagonal-star shaped template to difference sizes and then
check if some of the resized templates has a well matched patch of the
same size in the reference image \cite{hinterstoisser2012gradient}.
If the best matched patch can be any size from $\alpha n \times \alpha
n$ to $\beta n \times \beta n$, where $n$ is the side length of the
query template and $\alpha, \beta$ are some constants, then there are
$\beta n - \alpha n = O(n)$ different possible template scales to
examine.  In the case of our pre-processing algorithm, as comparing
the features of a template to all patches of the same size requires
$O(M^2+n^2)$ time for a $M \times M$ reference image, the overall time
complexity for scale invariant screening with the aforementioned
method is $O(M^2n + n^3)$.

The exhaustive search strategy is inefficient and unsuited for our
goal of fast screening of unmatched regions, especially when the query
template is large.  But this method can be greatly accelerated if we
only check patches of a few different sizes in the reference image
instead of enumerating every possible sizes.  The idea is based on an
assumption that, if two patches of the same size have matched central
area, then these two patches might also match.  For instance, given an
$n \times n$ query template $T$, suppose that image $Y_k$ of size $k
\times k$ is an up-scaled version of $T$ as shown in
Figure~\ref{fig:scale}, and the patch feature vector $f_{k,n}$ of the
$n \times n$ center of $Y_k$, namely $Z_{k,n}$, matches a patch $P_n$
in the reference image, then the $k \times k$ patch $P_k$ centered at
the same location as $P_n$ might also match the features of $Y_k$.
Similarly, given $Y_{k+1}$, a $(k+1) \times (k+1)$ version of $T$, if
it matches a patch $P_{k+1}$ of the same size, their $n \times n$
central areas, $Z_{k+1,n}$ and $P_n$, should have matching features.
Therefore, to find the matched patches of either size $k \times k$ or
$(k+1) \times (k+1)$, we only have to look through all the $n \times
n$ patches in the reference image; there is no need for calculating
the features for patches of different sizes.

The assumption that matched central areas indicate matched patches is
not always true obviously, however, it only slightly increases the
possibility of mistaking a wrong patch as matched, as long as the
examined center area is sufficiently large, i.e., $n$ is not much
smaller than $k$.  Moreover, the best match should have a matched
central area to the query template hence not affected by the
assumption.  Therefore, despite the limitation of the assumption, the
pre-processor can still rule out a great amount of unlikely matched
areas without removing the best match for the next stage template
matching algorithm.

Although this method can process the templates of several different
scales with only the patches of one fixed size, it still needs to
compare the features of a patch with the features of the query
template of every scale.  Thus, the time complexity for comparing all
the templates with all the patches is still $O(M^2n)$.  To make the
feature comparison function more efficient, we can aggregate the
features of the templates of different scales together into a feature
set $F_n$,
\begin{align}
  F_n = \{f_{k,n} \,|\, n \le k \le \lambda n \},
  \label{eq:F_n}
\end{align}
which includes all the feature vectors of the scaled templates of size
from $n \times n$ to $\lambda n \times \lambda n$.  Now given $F_n$,
we can examine if a $n \times n$ patch $P$ matches $Z_{k,n}$ for some
$k \in [n, \lambda n]$ by testing the membership of the feature vector
of $P$ in set $F_n$.  If the size of the best matched patch ranges
from $\alpha n \times \alpha n$ to $\beta n \times \beta n$, then we
need to repeat the process $\log_{\lambda}(\beta / \alpha)$ times for
different scale ranges.

The feature set $F_n$ can be implemented using a membership array
$\hat{F}_n$ with $\hat{F}_n(Q(f)) = 1$ for any $f \in F_n$, where
$Q(f)$ is a quantizer for feature vector.  For simplicity's sake,
assume that there is only one feature to consider and $Q(f) = \lfloor
f / q + \frac{1}{2} \rfloor$ is a uniform quantizer with quantization
factor $q$.  For some $f \in F_n$, if both $Q(f-q/2)$ and $Q(f+q/2)$
are marked as $1$ in $\hat{F}_n$ in addition to $Q(f)$, then, given a
feature $g$, $\hat{F}_n(Q(g)) = 1$ as long as $|f - g| < q/2$.  Thus,
quantization factor $q$ is a parameter that sets how similar the
features of two patches should be before counting them as matched.

\begin{algorithm}
  \caption{\label{alg:scale}Screening pre-processor for scale invariant template
    matching}

  \DontPrintSemicolon

  \KwData{$I$, reference image}
  \KwData{$T$, query template}
  \KwData{$[\alpha, \beta]$, scale range}
  \KwResult{$R$, set of possible matched patches}

  \Begin{
    $\lambda \leftarrow \sqrt{2}$ \;
    $n \leftarrow \operatorname{get\_width}(T)$ \;
    $R \leftarrow \varnothing$ \;
    $m \leftarrow \beta n$ \;
    \While{$m \ge \alpha n$}{
      $m \leftarrow m / \lambda$ \;
      $F_m \leftarrow \varnothing$ \;
      \For{$k \leftarrow m$ \KwTo $m\lambda$}{
        $Y_k \leftarrow \operatorname{scale}(T, k \times k)$ \;
        $Z_{k,m} \leftarrow \operatorname{crop\_center}(Y_k, m \times m)$ \;
        $f_{k,m} \leftarrow Q(\operatorname{features}(Z_{k,m}))$ \;
        $F_m \leftarrow F_m \cup \{f_{k,m}\}$ \;
      }
      \ForEach{$m \times m$ patch $P$ in $I$}{
        \If {$Q(\operatorname{features}(P)) \in F_m$}{
          $R = R \cup \{P\}$ \;
        }
      }
    }
  }

\end{algorithm}

This efficient scale invariant pre-processing algorithm is summarized
in Algorithm~\ref{alg:scale}.  Since a membership array only requires
constant time to initialize and constant time to access one element
regardless of the size of the array \cite{storer2001introduction}, the
time complexity for constructing $F_n$ is $O(n^3)$ and the time
complexity for testing the features of all the patches is $O(M^2)$.
Overall, the proposed scale invariant pre-processing algorithm
requires $O(M^2 +n^3)$ time.  Considering that the reference image is
generally much larger than the query template, the increase in the
asymptotic computational complexity from $O(M^2 +n^2)$ due to the
addition of scale invariant support is insignificant.


\begin{figure}
  \centering
  \begin{subfigure}{0.49\linewidth}
    \centering
    \includegraphics[width=\textwidth]{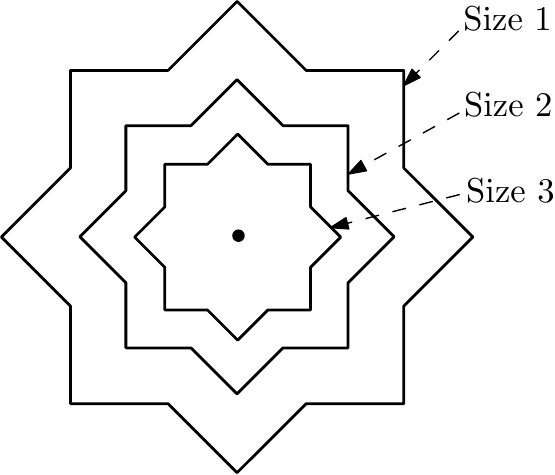}
    \caption{Central areas of different sizes}
    \label{fig:ring1}
  \end{subfigure} \quad
  \begin{subfigure}{0.42\linewidth}
    \centering
    \includegraphics[width=\textwidth]{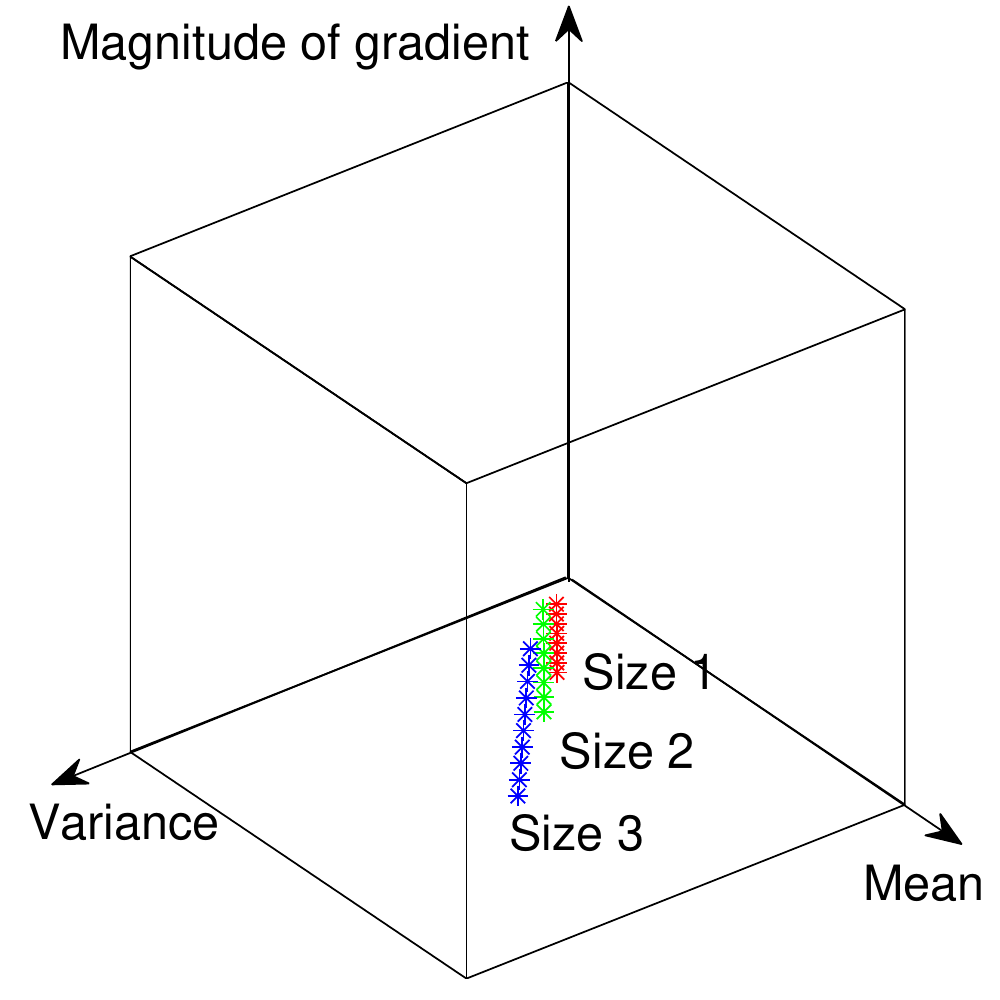}
    \caption{Vector set $F_n$ for central areas of different sizes}
    \label{fig:ring2}
  \end{subfigure}
  \caption{Features from different central areas of a patch are
    utilized together to improve matching accuracy.}
  \label{fig:ring}
\end{figure}

Patch features of different scales calculated for scale invariant
matching benefit the accuracy of matching as well.  As discussed
previously, we consider a $k \times k$ patch $P_k$ matches template
$Y_k$ of size $k \times k$ if their $n \times n$ central areas $P_n$
and $Z_{k,n}$ share the similar features.  In addition to $P_n$ and
$Z_{k,n}$, if the smaller $(n/\lambda) \times (n/\lambda)$ central
areas $P_{n/\lambda}$ and $Z_{k,n/\lambda}$ also have matched
features, patch $P_k$ is even more likely to be a match to the query
template.  Therefore, given a query template, the feature comparison
process should not be limited only to the large central area; the
additional features from smaller center areas, as illustrated in
Figure~\ref{fig:ring1}, can be used to provide extra details about a
patch, making the matching process more accurate and reliable.
Figure~\ref{fig:ring2} plots scale invariant feature vector set $F_n$
for each central area of a query template.  In our scale invariant
algorithm, if any one of three central areas of a patch does not match
the features of the corresponding areas of the template, the patch is
marked as unmatched.  Since these features of different scales
($n,n/\lambda,n /\lambda^2,\ldots$) have already been calculated for
scale invariant matching, the extra cost for utilizing these features
is negligible.

\section{Experimental Results}
\label{sec:result}


To evaluate the performance of the proposed technique, we implement
the rotation and scale invariant pre-processing algorithm, as shown in
Algorithm~\ref{alg:scale}, in C++.  Given a pair of reference image
and query template, the pre-processing program marks regions and
patches that may match the template and passes the results to a second
stage template matching algorithm to pinpoint the exact location of
the best match.  Three conventional high-precision template matching
techniques, SIFT \cite{lowe2004distinctive}, BBS \cite{dekel2015best}
and FAsT-Match \cite{korman2013fast}, are tested as the second stage
matching algorithm to demonstrate the effectiveness of the proposed
pre-processor for different types of template matching techniques.
The implementations of the three tested techniques are from their
original authors and executed with the default settings.  All of the
reported experiments in this paper are carried on a computer with an
Intel i7-4770 (3.4GHz) CPU and 8GB memory.

\subsection{Exp. I: Image Matching}


\begin{table}
  \centering
  \resizebox{\linewidth}{!}{
    \setlength\tabcolsep{1ex}
    \begin{tabular}{|c|cccccc|}
      \hline
      Data & Image & Template & Average & Pruning & Pruning & time \\
      set & size & size & overlap & patch & region & (s)\\
      \hline
      \hline
      I1; T1 & 640$\times$480 & 32$\times$32 & 99.53\% & 99.71\% & 81.36\% & 0.1 \\
      I1; T2 & 640$\times$480 & 64$\times$64 & 99.82\% & 99.72\% & 74.96\% & 0.1 \\
      I2; T1 & 960$\times$720 & 32$\times$32 & 99.28\% & 99.79\% & 88.87\% & 0.2 \\
      I2; T2 & 960$\times$720 & 64$\times$64 & 99.95\% & 99.84\% & 81.74\% & 0.2 \\
      I2; T3 & 960$\times$720 & 96$\times$96 & 99.92\% & 99.88\% & 74.29\% & 0.2 \\
      I3; T1 & 1280$\times$960 & 32$\times$32 & 99.94\% & 99.76\% & 85.87\% & 0.4 \\
      I3; T2 & 1280$\times$960 & 64$\times$64 & 99.98\% & 99.75\% & 80.83\% & 0.4 \\
      I3; T3 & 1280$\times$960 & 96$\times$96 & 99.99\% & 99.74\% & 77.64\% & 0.4 \\
      I3; T4 & 1280$\times$960 & 128$\times$128 & 99.97\% & 99.55\% & 72.63\% & 0.4 \\
      \hline
    \end{tabular}
  }
  \caption{Statistical results of each data set for image matching.}
  \label{tab:result}
\end{table}

In this group of experiments, we evaluate the performance of the
proposed pre-processing algorithm for high resolution images.  The
test images come from the MIT database \cite{torralba2009csail}, which
covers various scenes like urban streets, indoor and natural
environments.  We select all the 2250 grayscale images in the database
with a resolution no less than 640$\times$480 and divide them into
three data sets of different sizes with each set containing 750
images.  For each image, we extract 2 templates at random locations
with random rotations and then scale them to a given size.  The scale
range is $[0.5, 2]$, i.e., the scale parameter $\alpha, \beta$ in
Algorithm~\ref{alg:scale} are $0.5, 2$, respectively.  To guarantee
that these templates can be matched by the conventional template
matching techniques in the second stage, we require the standard
deviation of the pixel intensities of each template to be above a
threshold \cite{ouyang2012performance}.  As shown in
Table~\ref{tab:result}, 9 different combinations of image and template
sizes are used in our experiment.  Thus, there are 13500 test cases (9
data sets $\times$ 750 images $\times$ 2 templates) in total in this
experiment.

With regard to successful screening, we employ a similar definition as
in \cite{korman2013fast}: the regions preserved by the pre-processing
algorithm must overlap at least 90\% of the area of the ground truth,
otherwise, the screening is considered as failed.  By this definition,
our proposed scheme never fails in any of the 13500 test cases, and on
average, more than 99.8\% of the ground truth is preserved after the
screening, as presented in Table~\ref{tab:result}.  Since the matched
patch reported by a conventional template matching algorithm overlaps
with the ground truth only by 80\% on average in general, our scheme
has negligible impact on the accuracy of the second stage algorithm.


\begin{figure}
  \centering
  \begin{subfigure}{0.49\linewidth}
    \centering
    \includegraphics[width=\textwidth]{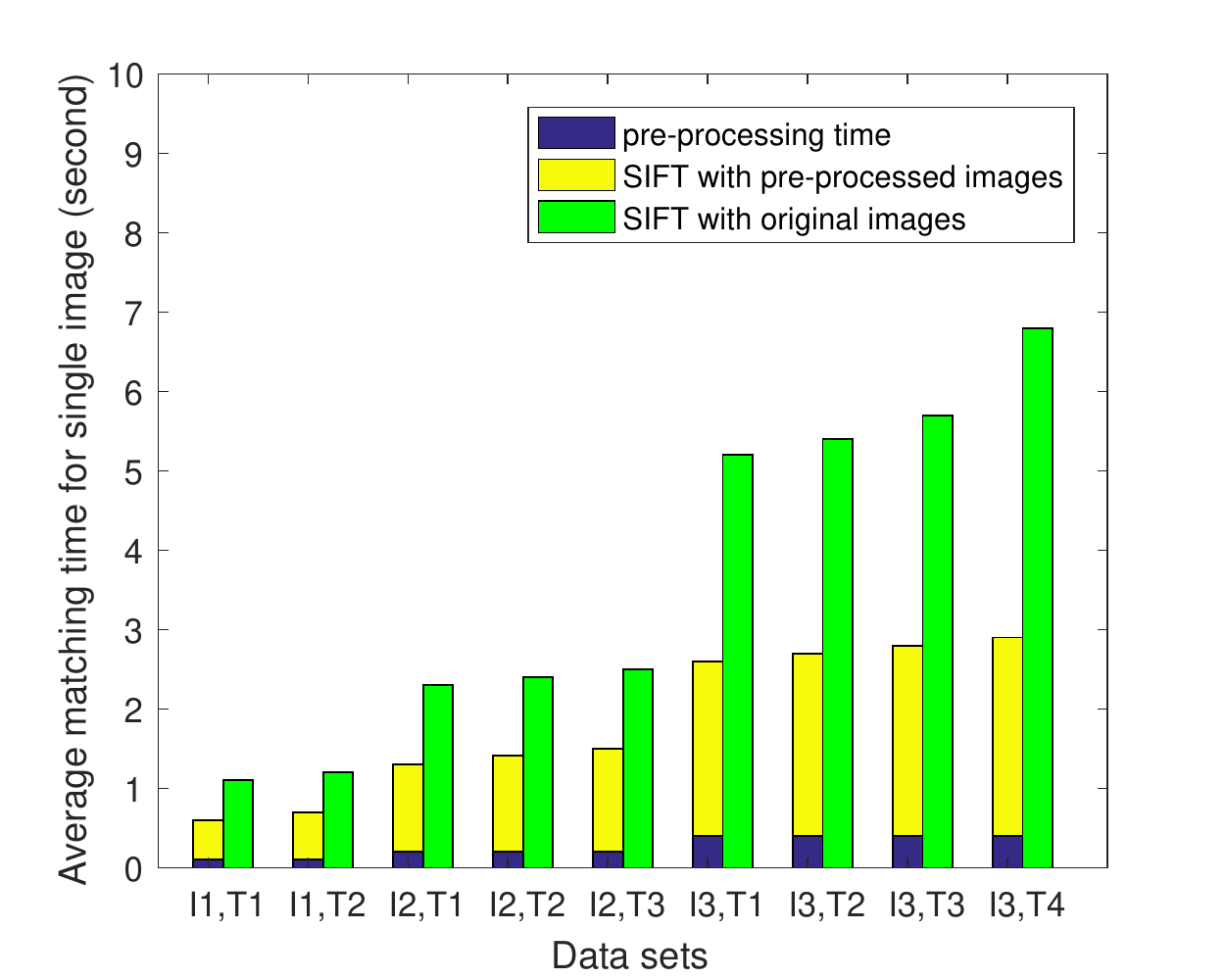}
    \caption{SIFT}
  \end{subfigure}
  \begin{subfigure}{0.49\linewidth}
    \centering
    \includegraphics[width=\textwidth]{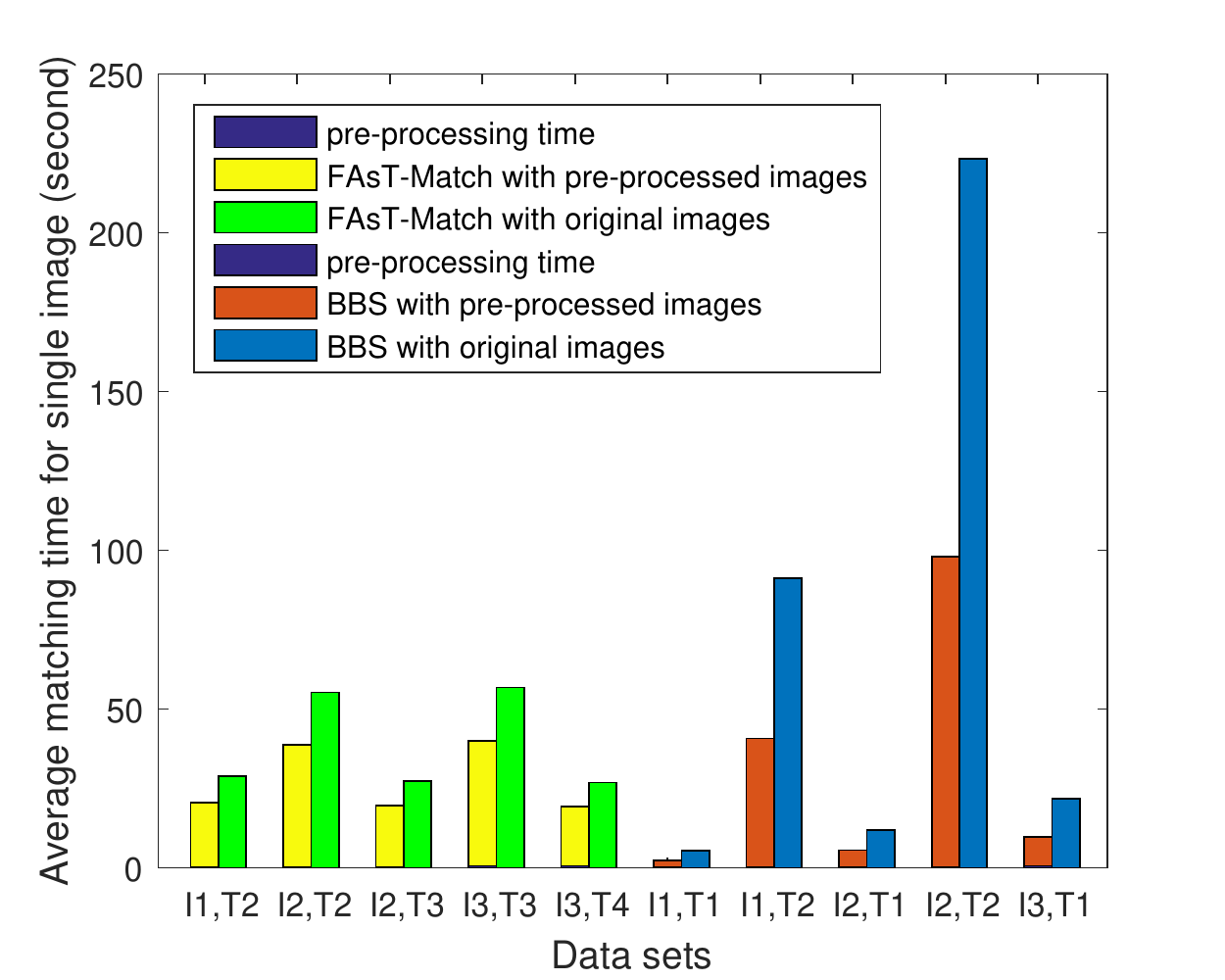}
    \caption{FAsT-Match and BBS}
  \end{subfigure}
  \caption{Comparison of the matching times with or without the
    proposed pre-processing algorithm.}
  \label{fig:time}
\end{figure}

\begin{figure}
  \centering
  \includegraphics[width=1\linewidth]{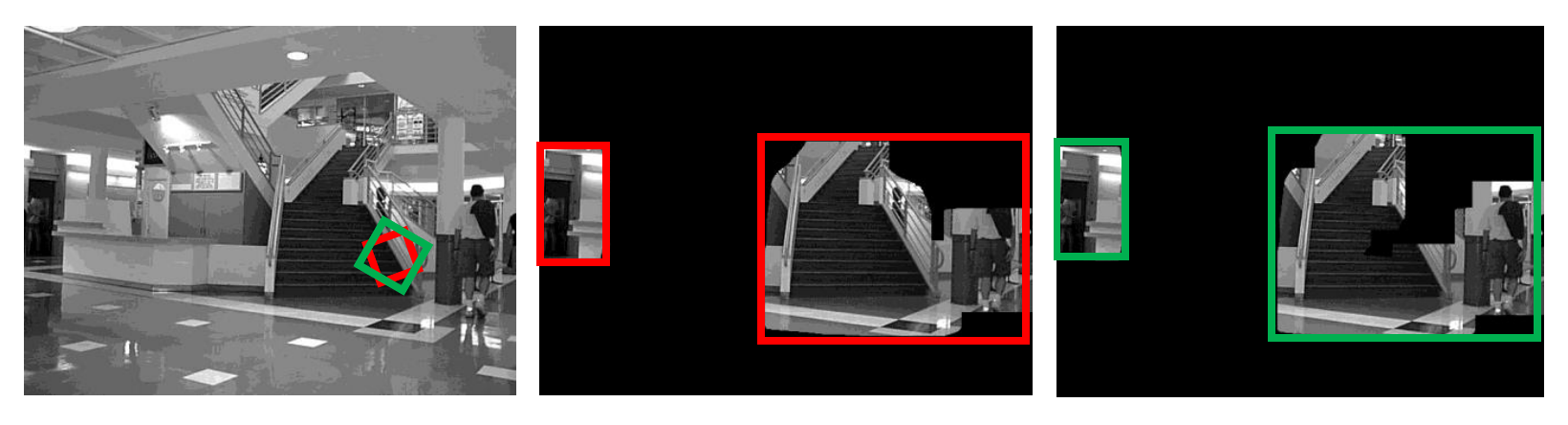}
  \includegraphics[width=1\linewidth]{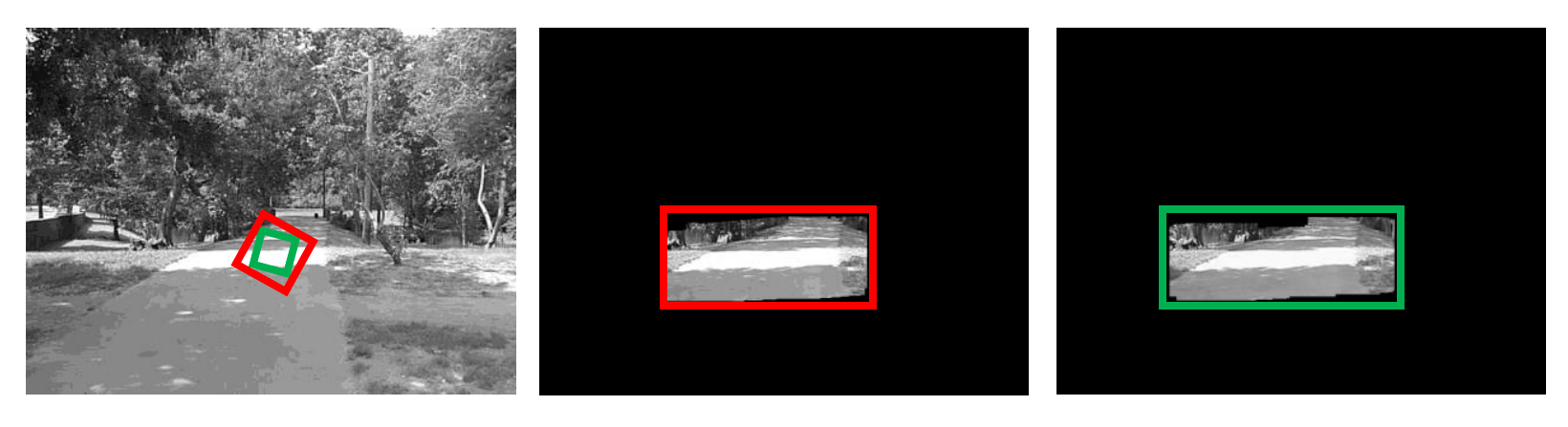}
  \includegraphics[width=1\linewidth]{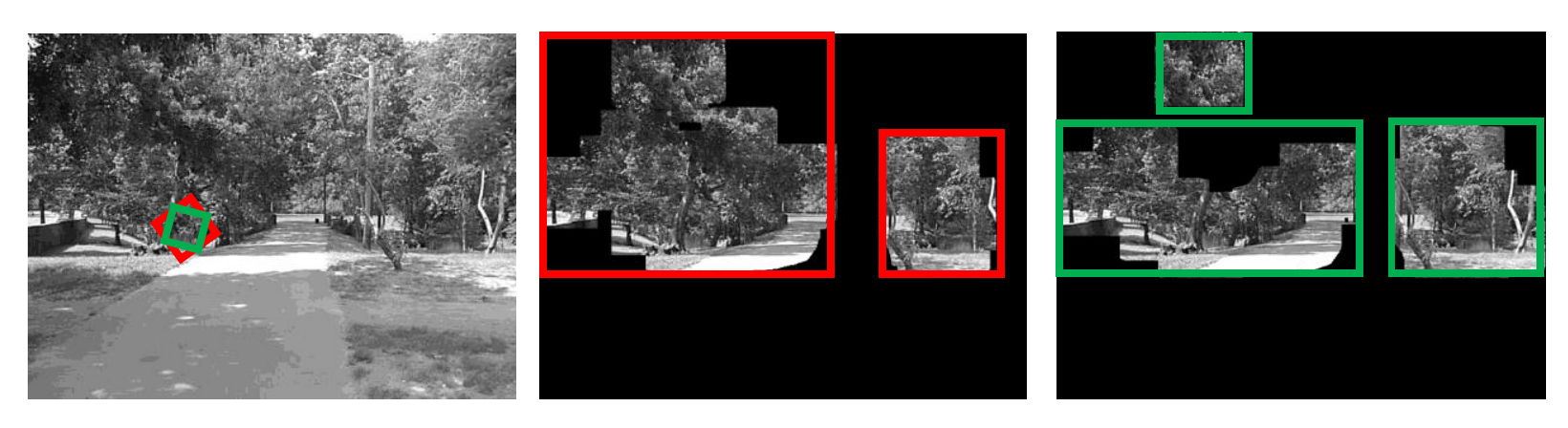}
  \includegraphics[width=1\linewidth]{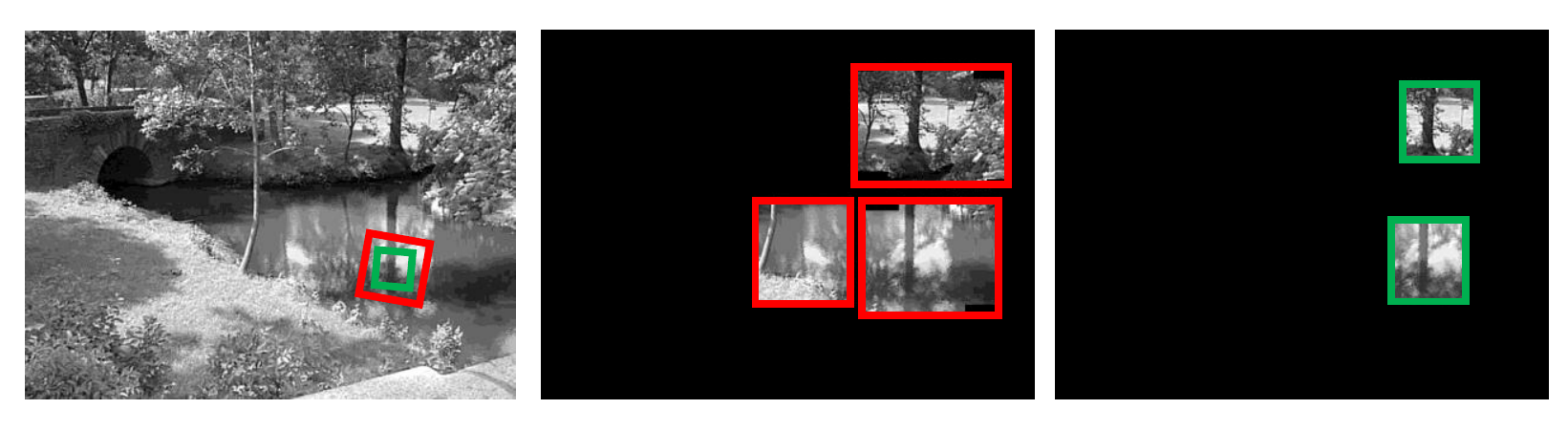}
  \caption{Example of screening results for image matching.}
  \label{fig:14}
\end{figure}

Furthermore, our scheme prunes around 80\% of the regions on average;
and only less than 0.5\% of all the candidate patches are marked as
possible matches and sent to the second stage algorithm.  As a result,
the second stage algorithm has much a smaller matching problem to
solve and hence requires shorter time.  For example, as shown in
Figure~\ref{fig:time}, SIFT uses 50\% less time by employing the
pre-processing algorithm and FAsT-Match and BBS also have 30\% and
55\% reductions, respectively, in time on average.  We also test BBS
with this data set, however, BBS cannot handle the large rotation and
scaling properly in the synthetic experiment, failing almost all the
test cases.  Figure~\ref{fig:14} shows some example results using the
MIT database.  The images in the left column are the reference images
with two query templates marked in boxes for each image.  The
remaining regions after screening are shown in the middle and right
columns.  In general, the more distinct the template is, the more
searching space can be ruled out by the proposed algorithm.

\begin{figure}
  \centering
  \begin{subfigure}{0.98\linewidth}
    \includegraphics[width=0.32\linewidth]{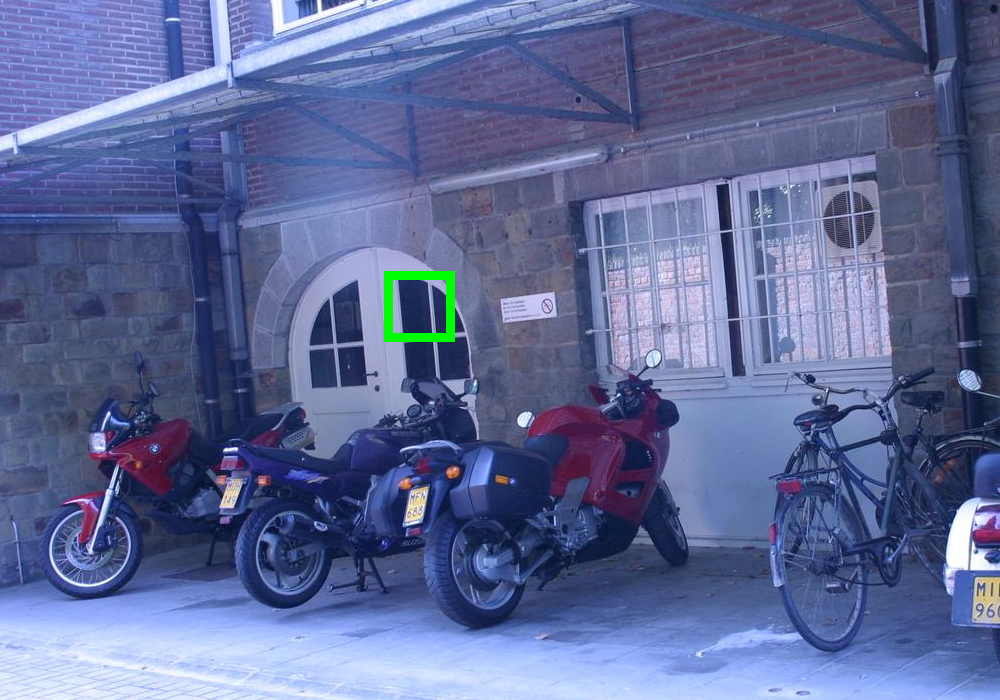}
    \includegraphics[width=0.32\linewidth]{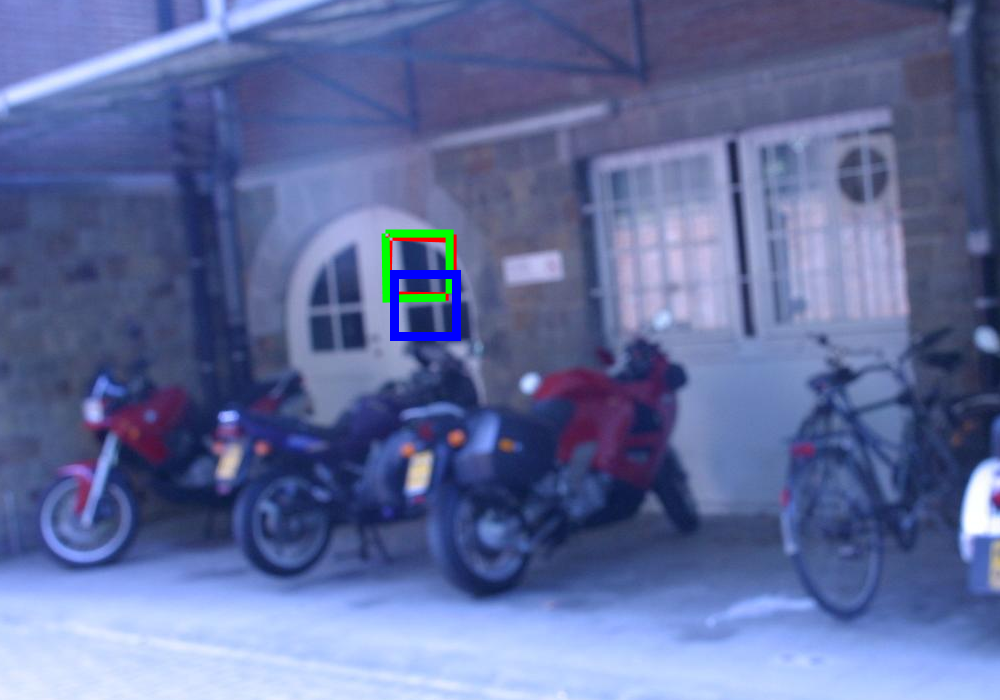}
    \includegraphics[width=0.32\linewidth]{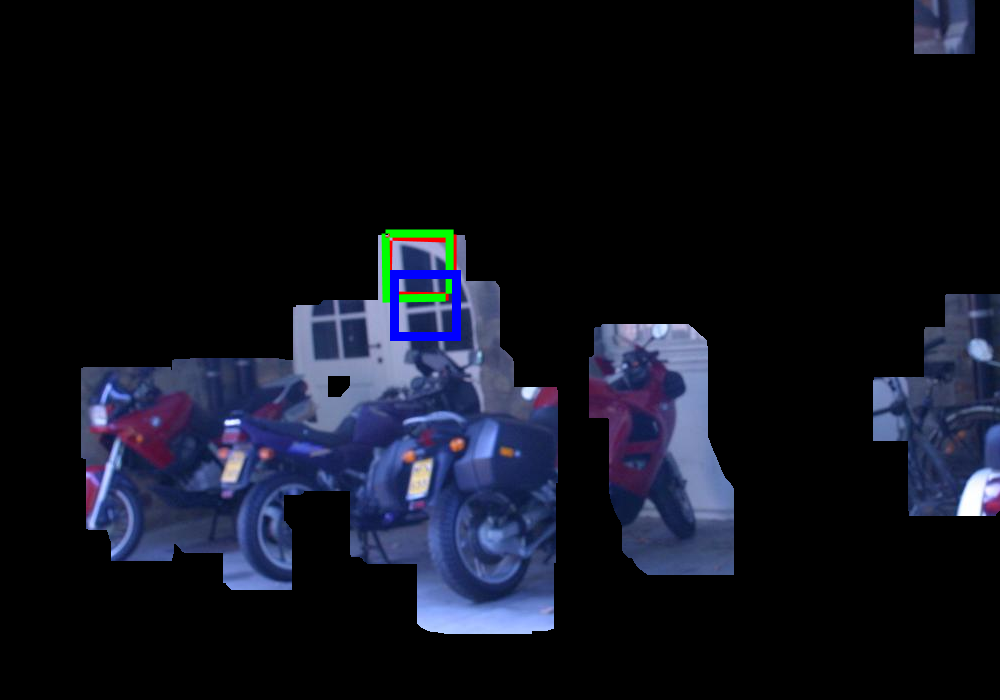}
    \vspace*{-1ex}
    \caption{Blur}
  \end{subfigure}
  \begin{subfigure}{0.98\linewidth}
    \includegraphics[width=0.32\linewidth]{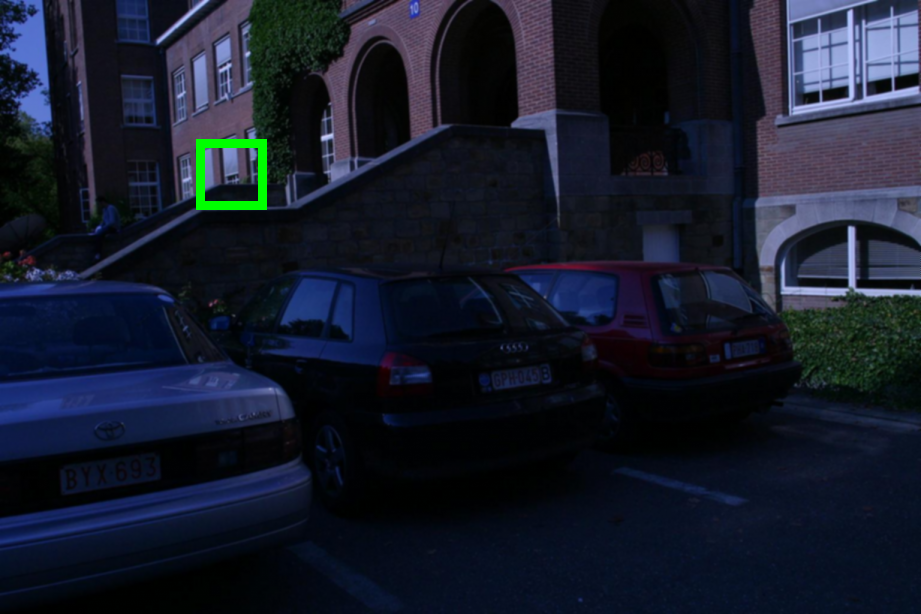}
    \includegraphics[width=0.32\linewidth]{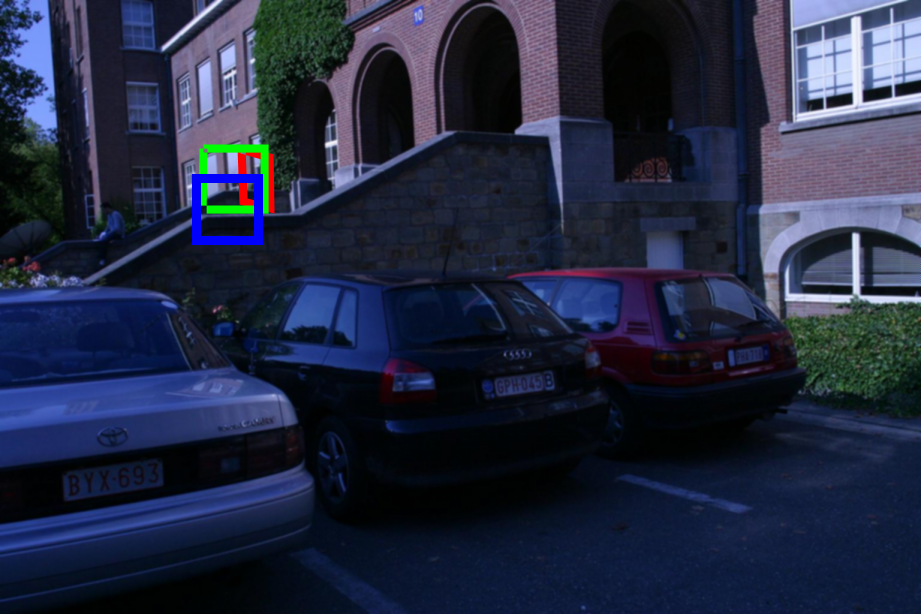}
    \includegraphics[width=0.32\linewidth]{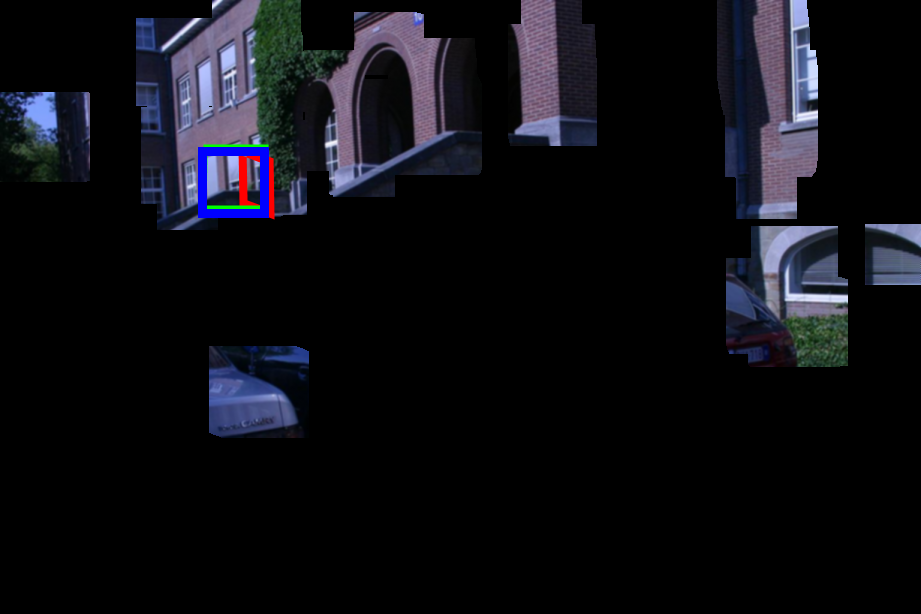}
    \vspace*{-1ex}
    \caption{Brightness change}
  \end{subfigure}
  \begin{subfigure}{0.98\linewidth}
    \includegraphics[width=0.32\linewidth]{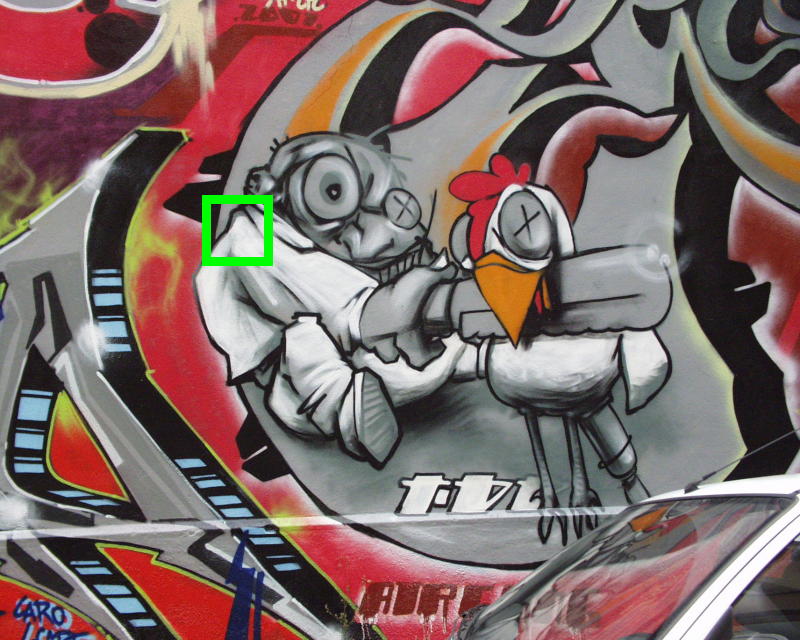}
    \includegraphics[width=0.32\linewidth]{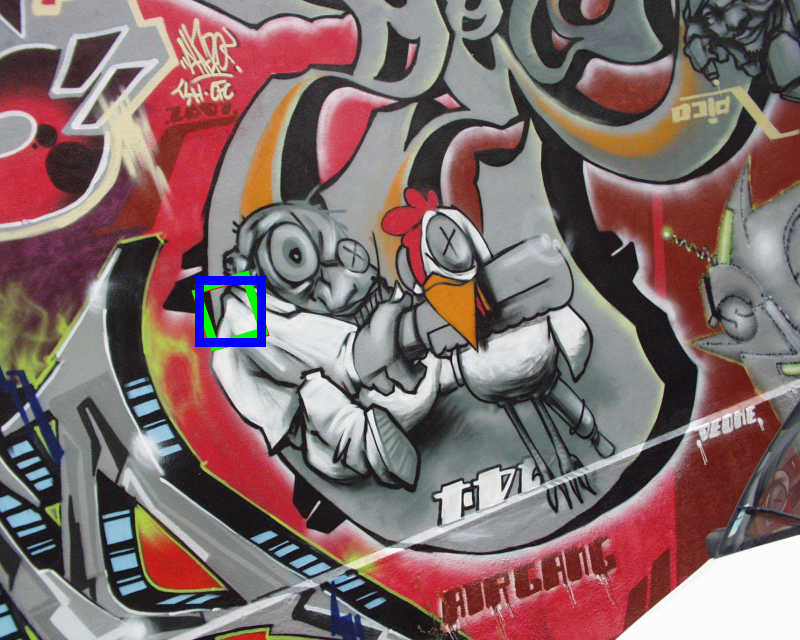}
    \includegraphics[width=0.32\linewidth]{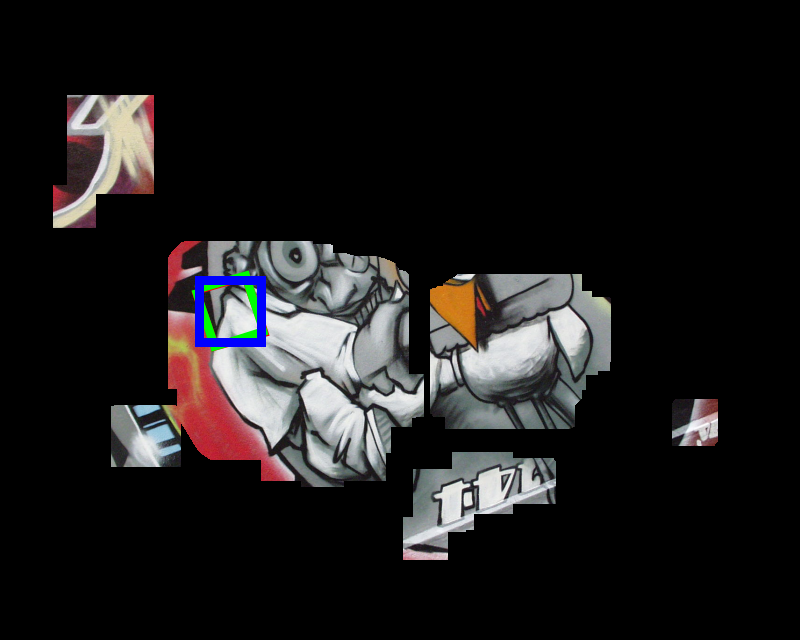}
    \vspace*{-1ex}
    \caption{Viewpoint change}
  \end{subfigure}
  \begin{subfigure}{0.98\linewidth}
    \includegraphics[width=0.32\linewidth]{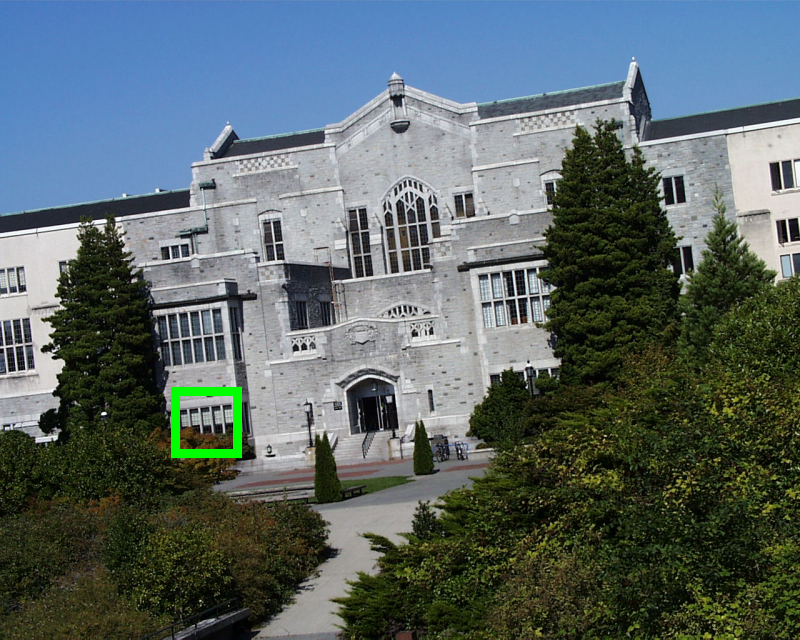}
    \includegraphics[width=0.32\linewidth]{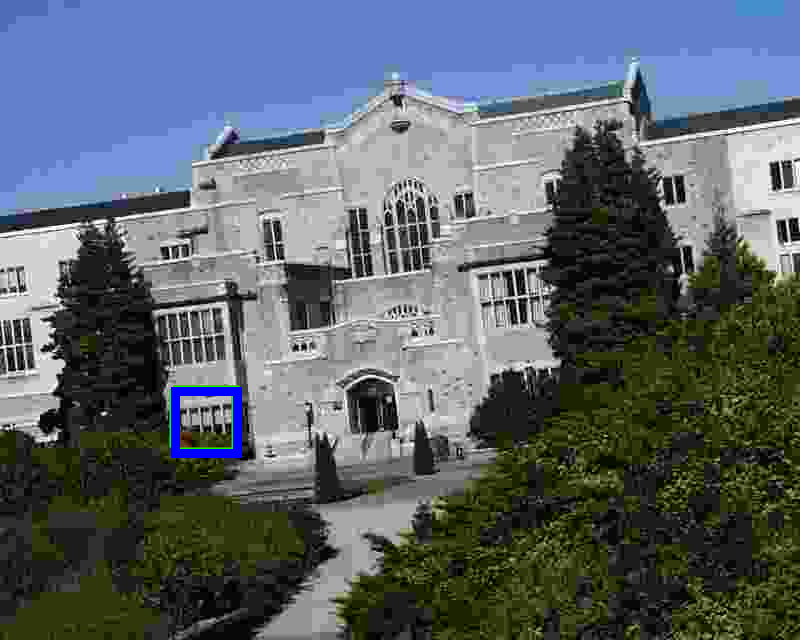}
    \includegraphics[width=0.32\linewidth]{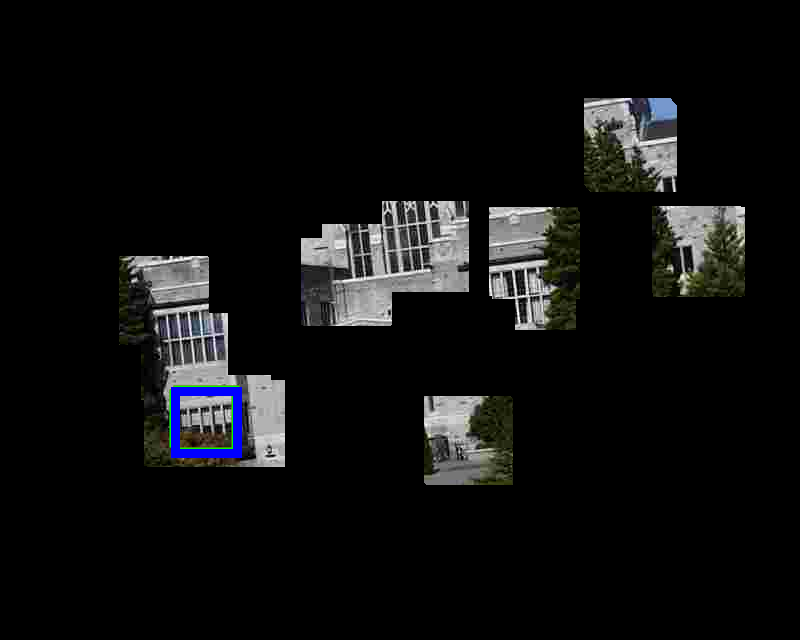}
    \vspace*{-1ex}
    \caption{JPEG compression}
  \end{subfigure}
  \caption{Example of screening results for images with distortions.}
  \label{fig:exp3}
\end{figure}

We also use the data set provided in \cite{korman2013fast,
  mikolajczyk2005performance}, to test our algorithm for images with
distortions, like blur, brightness change, viewpoint change, JPEG
compression.  Using the 90\% overlap criterion, our screening
algorithm successfully preserves the ground truth in all the cases
with blur, zoom, rotation and JPEG compression deformations.  But the
success ratio falls to around 60\% for test cases with significant
viewpoint or brightness change.  However, overall the screening has
negligible impact to the results of the conventional algorithms, as
they usually cannot find the best match as well for those difficult
cases.  As demonstrated in Figure~\ref{fig:exp3} are some sample
results for images with different distortions, where the green boxes
in the left column mark the templates and the green, red, blue boxes
in the middle and right columns mark the ground truth, the results by
FAsT-Match and the results by BBS, respectively.

\subsection{Exp. II: Video Tracking}

\begin{figure}
  \centering
  \includegraphics[width=0.32\linewidth]{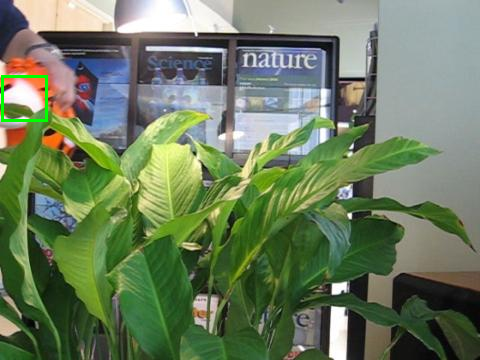}
  \includegraphics[width=0.32\linewidth]{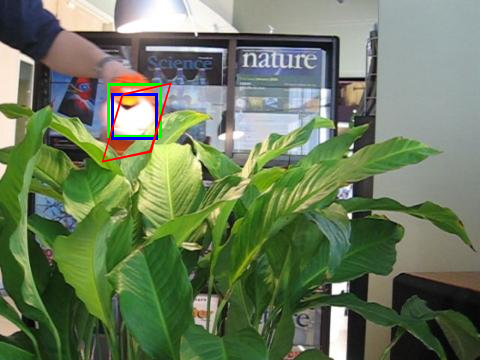}
  \includegraphics[width=0.32\linewidth]{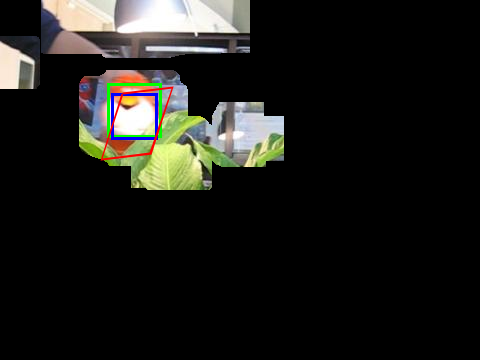}
  \\[0.1cm]
  \includegraphics[width=0.32\linewidth]{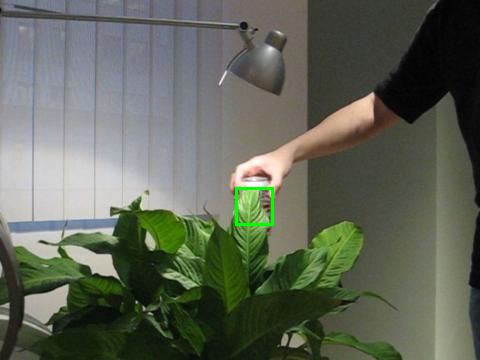}
  \includegraphics[width=0.32\linewidth]{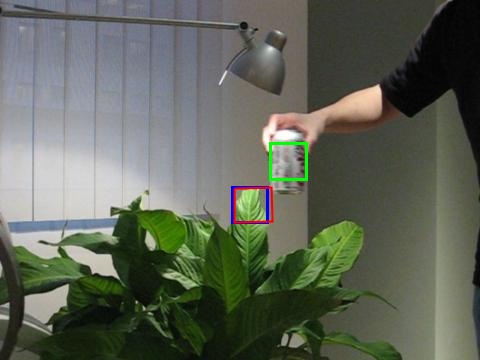}
  \includegraphics[width=0.32\linewidth]{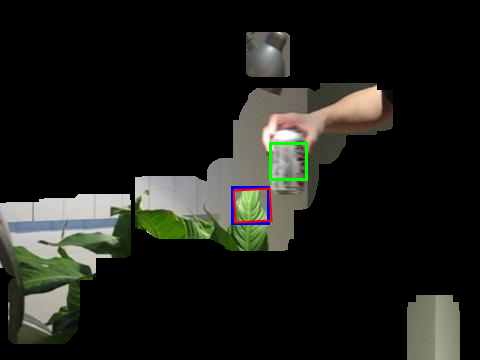}
  \\[0.1cm]
  \includegraphics[width=0.32\linewidth]{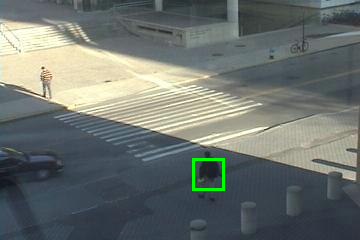}
  \includegraphics[width=0.32\linewidth]{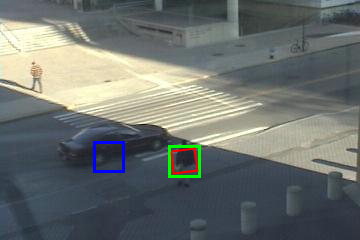}
  \includegraphics[width=0.32\linewidth]{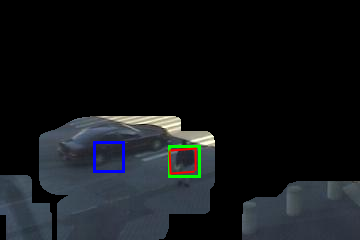}
  \\[0.1cm]
  \includegraphics[width=0.32\linewidth]{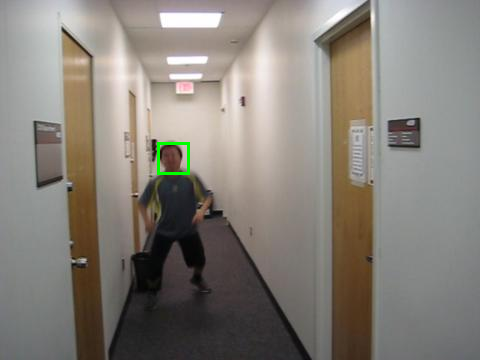}
  \includegraphics[width=0.32\linewidth]{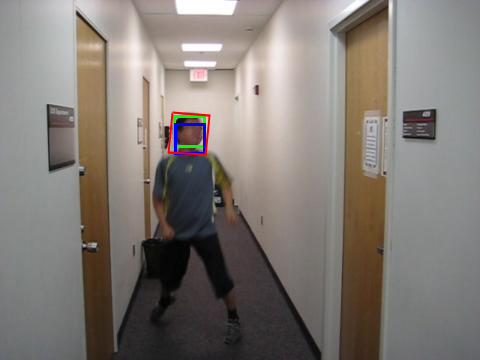}
  \includegraphics[width=0.32\linewidth]{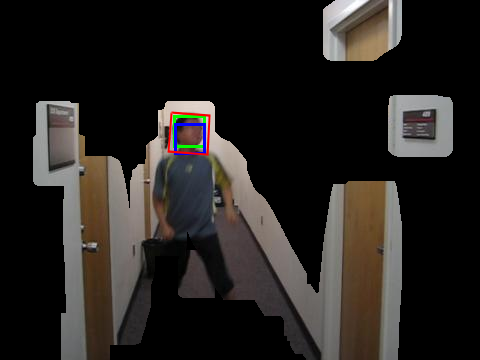}
  \caption{Example of screening results for video tracking.}
  \label{fig:video}
\end{figure}

This group of experiments evaluates our algorithm for video tracking
applications.  The test set is generated from 35 color video clips
provided in the Visual Tracker Benchmark \cite{wu2013online}.  From
each video, we randomly pick three pairs of frames that are 20 frames
apart and use the annotated object in the first frame as template and
the second frame as reference image for each frame pair.  This test
set is quite challenging, because the objects of interest typically
undergo some non-rigid deformations and may also be partially occluded
after 20 frames.

\begin{table}
  \centering
  \resizebox{\linewidth}{!}{
    \setlength\tabcolsep{1ex}
    \begin{tabular}{|c|c|c|c|}
      \hline
      Algorithms & Results & Without screening & With screening \\
      \hline
      \hline
      \multirow{3}{*}{FAsT-Match} & Time per image (s) & 36.2s & 25.3s \\
      & Average overlap & 41\% & 41\% \\
      & Success ratio & 45/105 & 46/105 \\
      \hline
      \multirow{3}{*}{BBS} & Time per image (s) & 16.7s & 9.4s \\
      & Average overlap & 62\% & 61\% \\
      & Success ratio & 75/105 & 75/105 \\
      \hline
      \multirow{3}{*}{SIFT} & Time per image (s) & 1.8s & 1.4s \\
      & Average overlap & N/A & N/A \\
      & Success ratio & 10/105 & 10/105 \\
      \hline
    \end{tabular}
  }
  \caption{Statistical results of each tested algorithm for video tracking.}
  \label{tab:result2}
\end{table}

The average running time of our pre-processing algorithm on this data
set is only 0.05s, as the resolution of the videos is relatively low
(480$\times$320).  On average, our algorithm prunes 88\% of the
patches and 55\% of the regions, and in 95 out of the 105 cases, it
preserves more than 90\% of the ground truth after screening.
Although in some cases, the screening is not successful ($>$90\%
overlap), our proposed algorithm does not adversely affect the success
ratio of the tested conventional algorithm at all as shown in
Table~\ref{tab:result2}.  Interestingly, the success ratio becomes
slightly higher for FAsT-Match when the screening algorithm is
applied.  This is because a conventional algorithm may match to a
wrong patch in some cases, but the wrong patch can be recognized and
removed by our algorithm beforehand.  Some sample results of the video
tracking data set are shown in Figure~\ref{fig:video}.  The results of
each algorithm are marked in the same way as Figure~\ref{fig:exp3}.

\section{Conclusion}
\label{sec:conclusion}

This paper presents a generic template matching pre-processor for
expediting conventional template matching techniques.  The proposed
pre-processing algorithm can handle arbitrary rotation and scaling of
reference images effectively as demonstrated by extensive experiments.

{\small
\bibliographystyle{ieee}
\bibliography{tmpmatch}
}

\end{document}